\documentclass[11pt]{article}
\usepackage{amsmath}
\usepackage{amssymb}

\usepackage[preprint]{acl}

\usepackage{times}
\usepackage{latexsym}
\usepackage{amsmath, amsthm, amsfonts}
\usepackage[T1]{fontenc}
\usepackage{enumitem}
\usepackage[utf8]{inputenc}

\usepackage{microtype}

\usepackage{inconsolata}

\usepackage{graphicx}

\usepackage[utf8]{inputenc}
\usepackage{newunicodechar}
\newunicodechar{σ}{\ensuremath{\sigma}}
\usepackage{booktabs}  % 用于绘制漂亮的表格
\usepackage{multirow}  % 用于合并单元格
\usepackage{pifont}    % 用于插入特殊符号
\usepackage{arydshln} % 用于虚线
\usepackage[vlined,ruled,boxed,commentsnumbered,linesnumbered]{algorithm2e}
\usepackage{pifont} % \ding{xx}
\usepackage{graphicx}
\usepackage{subcaption}
\usepackage{balance}
\usepackage[most]{tcolorbox} % 用于彩色文本框
\usepackage{xcolor}          % 颜色支持
\usepackage{graphicx}        % 插入图片
\usepackage{caption}         % 图片标题
\usepackage{multirow}
\usepackage{url}
\usepackage{tikz}
\usepackage{listings} % 插入代码用到
\usepackage{threeparttable}
\usepackage{listings}
\usepackage{color}
\lstdefinelanguage{json}{
  basicstyle=\ttfamily\small,
  showstringspaces=false,
  breaklines=true,
  colback=gray!25,
  string=[s]{"}{"},
  literate=
   *{:}{{{\color{black}{:}}}}{1}
    {,}{{{\color{black}{,}}}}{1}
    {"conversations"}{{{\color{purple!60!black}\texttt{"conversations"}}}}{15}
    {"from"}{{{\color{violet}\texttt{"from"}}}}{6}
    {"value"}{{{\color{teal!70!black}\texttt{"value"}}}}{7}
    {"human"}{{{\color{blue!70!black}\texttt{"human"}}}}{7}
    {"gpt"}{{{\color{blue!70!black}\texttt{"gpt"}}}}{5}
    {"true"}{{{\color{orange!80!black}\texttt{"true"}}}}{6}
    {"false"}{{{\color{orange!80!black}\texttt{"false"}}}}{7}
    {"null"}{{{\color{gray!70!black}\texttt{"null"}}}}{6}
    {"system"}{{{\color{purple!60!black}\texttt{"system"}}}}{15}
}

\definecolor{dkgreen}{rgb}{0,0.6,0}
\definecolor{gray}{rgb}{0.5,0.5,0.5}
\definecolor{mauve}{rgb}{0.58,0,0.82}
\usepackage{xcolor}
\usepackage{listings}

\title{ChatAD: Reasoning-Enhanced Time-Series Anomaly Detection with Multi-Turn Instruction Evolution}

\author{
  \textbf{Hui Sun\textsuperscript{1\thanks{Work done during an internship at MSRA.}}},
  \textbf{Chang Xu\textsuperscript{2\thanks{Corresponding author}}},
  \textbf{Haonan Xie\textsuperscript{3}},
  \textbf{Hao Li\textsuperscript{4}},
  \textbf{Yuhao Huang\textsuperscript{5}}, \\
  \textbf{Chuheng Zhang\textsuperscript{2}},
  \textbf{Ming Jin\textsuperscript{6}},
  \textbf{Xiaoguang Liu\textsuperscript{1}},
  \textbf{Gang Wang\textsuperscript{1}},
  \textbf{Jiang Bian\textsuperscript{2}},
\\
  \textsuperscript{1} Nankai University,
  \textsuperscript{2} Microsoft Research Asia, 
  \textsuperscript{3} Huanjiang Laboratory, \\
  \textsuperscript{4} University of Manchester, 
  \textsuperscript{5} Nanjing University, 
  \textsuperscript{6} Griffith University
\\
}
\usepackage{titlesec}

\titleformat{\paragraph}
    {\normalfont\normalsize\bfseries} % 字体样式：标准大小，加粗
    {\theparagraph}                  % 标题编号
    {1em}                            % 编号与标题间的距离
    {}                               % 标题前的特殊代码
    
\titlespacing*{\paragraph}
    {0pt}{3.25ex plus 1ex minus .2ex}{1.5ex plus .2ex}
\begin{document}
\maketitle
\begin{abstract}
LLM-driven Anomaly Detection (AD) helps enhance the understanding and explanatory abilities of anomalous behaviors in Time Series (TS). Existing methods face challenges of inadequate reasoning ability, deficient multi-turn dialogue capability, and narrow generalization. To this end, we 1) propose a multi-agent-based \underline{TS} \underline{Evol}ution algorithm named \textbf{TSEvol}. On top of it, we 2) introduce the AD reasoning \& multi-turn dialogue \underline{Data}set \textbf{TSEData-20K} and contribute the \underline{Chat}bot family for \underline{AD}, including \textbf{ChatAD}-Llama3-8B, $\sim$Qwen2.5-7B, and $\sim$Mistral-7B. Furthermore, 3) we propose the \underline{TS} \underline{K}ahneman-\underline{T}versky \underline{O}ptimization (TKTO) to enhance ChatAD's cross-task generalization capability.
Lastly, 4) we propose a \underline{L}LM-driven \underline{L}earning-based \underline{AD} \underline{Bench}mark \textbf{LLADBench} to evaluate the performance of ChatAD and nine baselines across seven datasets and tasks. Our three ChatAD models achieve substantial gains, up to 34.50\% in accuracy, 34.71\% in F1, and a 37.42\% reduction in false positives. Besides, via KTKO, our optimized ChatAD achieves competitive performance in reasoning and cross‑task generalization on classification, forecasting, and imputation. %The resources will be released upon acceptance of the paper.
\end{abstract}

\section{Introduction}

Learning-based Anomaly Detection Algorithms (LADAs) for Time Series (TS) demonstrates competitive advantages in improving efficiency and reducing risk like financial fraud and healthcare~\cite{timemqa}. Existing LADAs are divided into traditional Neural Network (NN)-based and LLM-driven. 
Among them, NN-based LADAs mine anomaly patterns from raw TS via deep modeling such as local and global dependency~\cite{adllm}. As shown in Fig.~\ref{intro}, compared with LLM-driven methods, NN-based LADAs feature a lightweight design but exhibit lower generalization and multimodal processing capabilities.

{\small
% \begin{figure*}[!thbp]
\begin{figure}[t!]
  \centering
  \includegraphics[width=1\columnwidth]{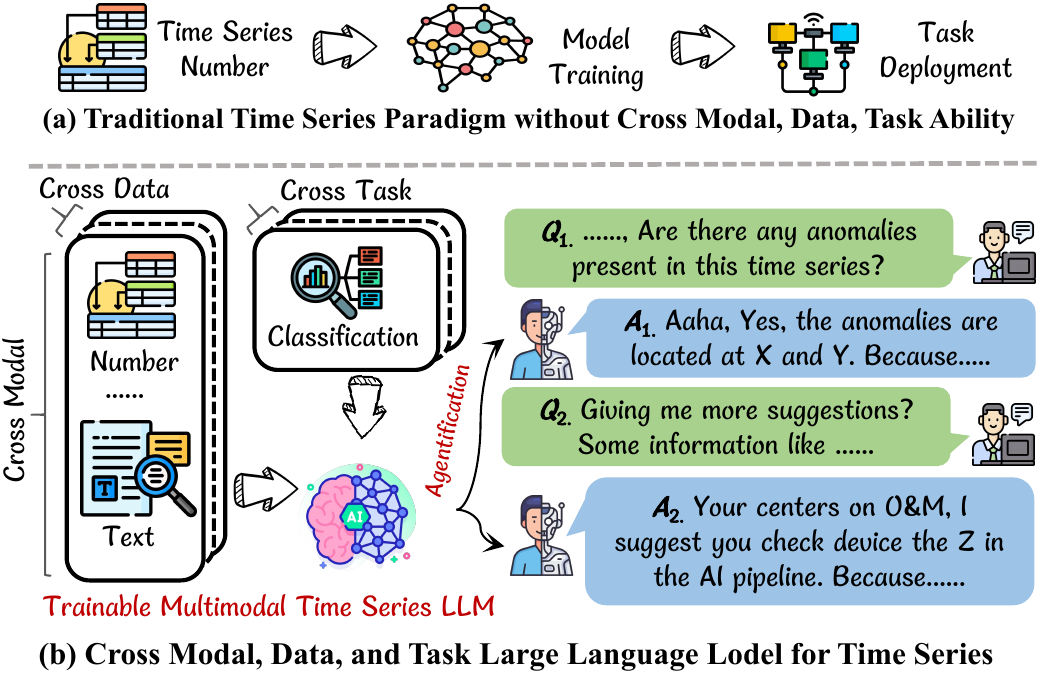}
  \caption{Comparison of modeling paradigms for time series. (a) Traditional NN-based methods. (b) Enhanced paradigm using MTSLLM as an universal generator.}
  \label{intro}
\end{figure}
}

Recently, Multimodal TS LLMs (MTSLLMs) have made progress in TS forecasting, classification, and generation~\cite{deng2025tardiff,kostina2025large}.
By integrating multimodal data, some early LLM-driven methods have enhanced the accuracy of AD task~\cite{liu2024large,liu2025large}. Subsequently, TimeMQA~\cite{timemqa} and TimeRA~\cite{timera} convert the AD classification problem into a generative task by designing Supervised Fine-Tuning (SFT) datasets. By distilling inherent patterns from SFT datasets, those MTSLLMs acquire fundamental TS analytical capabilities. However, to the best of our knowledge, these LLM-driven approaches still exhibit several inherent limitations and technical challenges.

Firstly, LLM-driven MTSLLMs suffer from \textit{\textbf{limited reasoning proficiency}}. Anomalies often manifest as intricate and latent patterns. Robust reasoning is essential for models to decipher these non-obvious correlations and perform causal analysis within TS data. This deficiency stems from: 1) the insufficient diversity and complexity of existing training corpora, and 2) the absence of intrinsic logical or causal reasoning mechanisms.

Secondly, current MTSLLMs exhibit a \textit{\textbf{deficient capacity for multi-turn dialogue interactions}}. Such interactions are vital for enabling users to conduct in-depth diagnostic inquiries into anomalies, thereby broadening cross-domain utility and facilitating informed decision-making. Empirically, we observed that enhancing multi-turn dialogue elevated the accuracy of Mistral-7B from 51.53\% to 70.17\%. Despite these advantages, existing models remain constrained by their reliance on single-turn SFT datasets, which results in suboptimal performance in complex, multi-turn TS Q\&A scenarios.

Thirdly, existing models exhibit \textit{\textbf{narrow generalization capabilities}}, manifest in two dimensions. 1) Task Level: The cross-task generalization of models like TimeMQA~\cite{timemqa} relies heavily on fine-tuning within specific domains; consequently, knowledge acquired from a single time-series task cannot be effectively transferred to others. 2) Data Level: Constrained by the limited inferential efficiency of MTSLLMs, current models often rely on a simplified "compressed memory" of fine-tuning datasets, which hampers their ability to generalize to unseen data distributions.

Strengthening AD comprehension and multi-turn dialogue within MTSLLMs paves the way toward fully explainable anomalies. Furthermore, superior generalization is essential for minimizing operational overhead and scaling real-world deployments. Driven by these goals, we present our contributions in the following:

\textbf{1)} We propose \textbf{TSEvol}, an LLM-driven agent-based time series evolution algorithm for SFT instruction synthesis. TSEvol features two novel role-play layers: a Cognitive Reasoning Layer (CRL) that mimics a psychological diagnosis (Consultant-Client) to construct CoT-based sequential reasoning data, and an Interactive Feedback-driven Layer (IFL) that simulates a Supervisor-Intern diagnostic scenario to enhance multi-turn dialogue quality.

\textbf{2)} We present the first LAD reasoning and multi-turn dialogue dataset \textbf{TSEData-20K} using TSEvol, comprising 10011 high-quality anomaly detection dialogues using the state-of-the-art GPT-5~\cite{openai_gpt5} model, spanning 8 domains with a total of 21303 dialogue turns. Building upon TSEData-20K, we contribute \textbf{ChatAD}, the first multi-turn dialogue and reasoning MTSLLM family for LAD, including three foundation model ChatAD-Llama3-8B,  $\sim$Qwen2.5-7B, and $\sim$Mistral-7B. 

\textbf{3)} We develop \textbf{TKTO} (\underline{T}ime \underline{S}eries \underline{K}ahneman-\underline{T}versky \underline{O}ptimization), a reinforcement learning approach designed to refine ChatAD. By training on minimal cross-task data, TKTO empowers the model with superior reasoning and zero-shot cross-task generalization.

\textbf{4)} We establish \textbf{LLADBench} to evaluate the ChatAD against nine state-of-the-art baselines. Extensive experiments demonstrate that ChatAD models achieve substantial performance gains: in single-turn scenarios, average accuracy and F1 scores increase by up to 34.50\% and 34.71\%, while FPR is reduced by up to 37.42\%. In multi-turn dialogues, ChatAD consistently outperforms baselines, with a remarkable FPR reduction of up to 52.44\%. These results, coupled with robust data- and task-level evaluations, confirm ChatAD's superior reasoning and generalization capabilities.

\section{Problem \& Tasks Definition}~\label{background}
Let $H = \{h_0, h_1, \dots, h_{n-1}\}$ denote a TS data of length $n$, where $h_i \in \mathbb{R}^{j \times k}$ represents $j$ channels with $k$ dimensions. Let $B$ and $\theta$ represent textual description and learnable parameters. We formulate LLM-driven LAD as a multimodal Question ($\mathcal{Q}$) and Answer ($\mathcal{A}$) generation task:
\begin{equation}
\label{eq:lad}
\mathcal{A} = \text{MTSLLM}_{\theta} (\mathcal{Q}(H, B))
\end{equation}
While existing MTSLLMs~\cite{timemqa,timera,huang2025deep} predominantly rely on single-turn Q\&A, our proposed \textbf{ChatAD} extends this paradigm. By leveraging the TSEvol and reinforcement learning, ChatAD enhances AD solutions across four critical dimensions: detection effectiveness, reasoning depth, multi-turn interaction, and cross-task/cross-dataset generalization.

{\small
% \begin{figure*}[!thbp]
\begin{figure*}[t]
  \centering
  \includegraphics[width=1\textwidth]{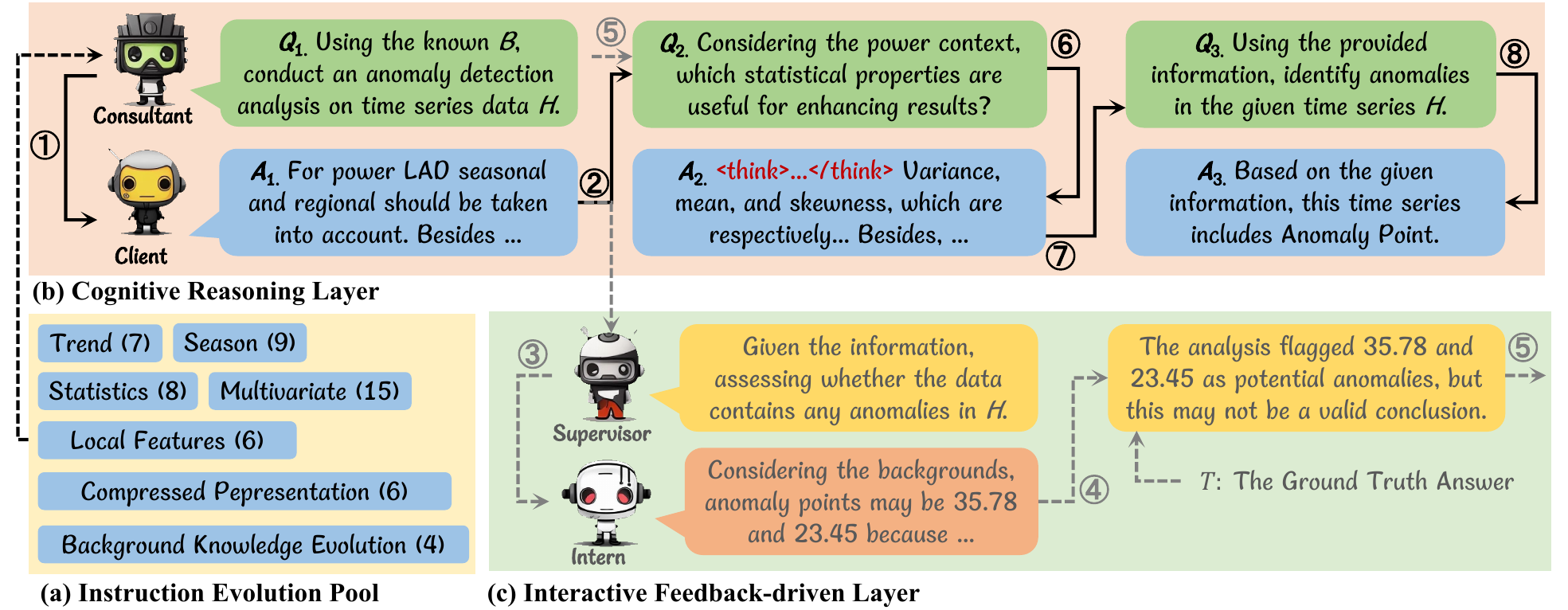}
  \caption{Pipeline of TSEvol (Only the evolution of $A_1$ and $Q_2$ are shown, with respect to the Interactive Feedback-driven Layer). Here, $H$ and $B$ denote the inputted history TS data, and short description of the TS backgrounds, respectively. The arrow $\rightarrow$ denotes information flow, with only the solid elements retained as the final result.}
  \label{tsevol}
\end{figure*}
}
\section{Related Work} \label{reletated_work}
\subsection{LLM-driven Anomaly Detection} LLM-driven algorithms reformulate AD as a generative task, offering superior cross-dataset generalization and multimodal integration compared to traditional numerical modeling paradigm. Representative works include LogiCode~\cite{LogiCode}, AadLLM~\cite{AadLLM}, and AnomalyLLM~\cite{AnomalyLLM}, et al. Recently, inspired by the success of reasoning-based LLMs~\cite{liu2024deepseek}, there has been a burgeoning interest in leveraging data augmentation to bolster the reasoning and conversational capabilities of MTSLLMs. Notably, TimeMQA~\cite{timemqa} pioneered cross-task MTSLLMs by constructing specialized dialogue datasets for single-turn reasoning. Subsequently, TimeRA~\cite{timera} introduced the RATs40K dataset to enhance MTSLLM alignment.
Despite these advancements, existing methods remain constrained by a scarcity of high-quality, complex training data. This results in significant performance bottlenecks in deep TS reasoning, sustained multi-turn dialogue, and robust generalization across unseen domains.

\subsection{Instruction Data Evolution (IDE)} 
IDE enhances LLM capabilities by iteratively generating and optimizing SFT data across text-based and multimodal domains. Frameworks such as Evol-Instruct~\cite{WizardLM} and Review-Instruct~\cite{Review-Instruct} progressively refine linguistic complexity through in-depth evolution and multi-agent review. Similarly, multimodal approaches like MMEvol~\cite{Mmevol} and LLaVA-Instruct~\cite{liu2023visual} leverage advanced models (e.g., GPT-4V) to synthesize high-quality image-text SFT data via fine-grained perception and cognitive reasoning.
Despite their success, existing IDE methods are ill-suited for TS, as they fail to account for domain-specific characteristics like seasonality and statistical properties.
{\small
% \begin{figure*}[!thbp]
\begin{figure*}[t]
  \centering
  \includegraphics[width=1\textwidth]{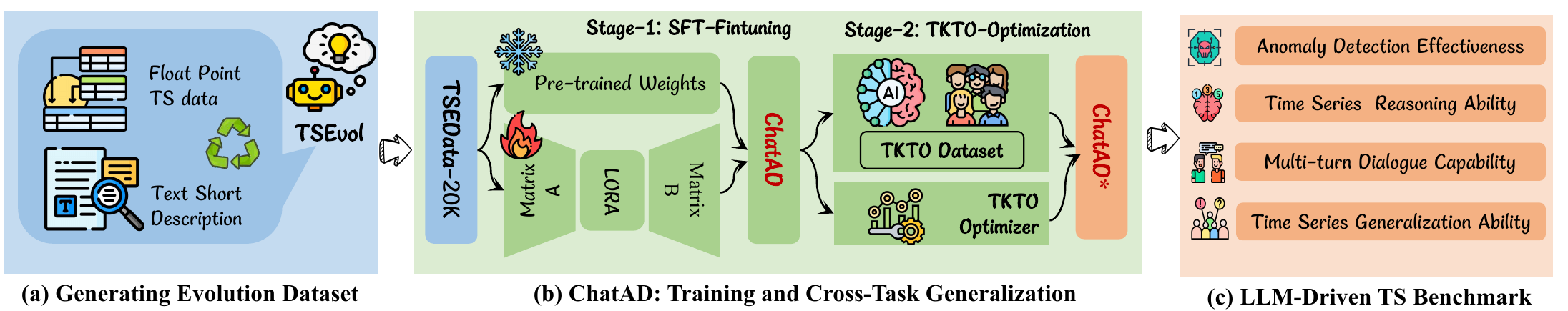}
  \caption{The model training and benchmark pipeline of time series data anomaly detection (ChatAD) model.}
  \label{mtsllm}
\end{figure*}
}
\subsection{Reinforcement Learning for LLM}
To bridge the gap between model outputs and complex reasoning, Reinforcement Learning (RL) and preference optimization have become pivotal. PPO~\cite{ppo} pioneered online alignment but remains computationally intensive, leading to the development of GRPO~\cite{deepseekv3}, which optimizes efficiency via group-relative rewards. To simplify this pipeline, DPO~\cite{dpo} enables direct optimization on preference pairs without an explicit reward model, while GAPO~\cite{gapo} improves stability through anchor-guided learning. Notably, KTO~\cite{kto} relaxes data constraints by utilizing unpaired binary signals. Given the inherent difficulty in acquiring paired TS labels, KTO's reliance on unpaired data offers a distinct advantage for cross-task optimization of MTSLLMs.

\section{Methods}\label{design}
This section first introduces the proposed multiple agent-based instruction data evolution algorithm, followed by the ChatAD fine-tuning pipeline, and finally the LLM-driven LAD benchmark.

\subsection{Time Series Instruction Data Evolution }
As shown in Fig.~\ref{tsevol}, TSEvol consists of (a) TS Instruction Evolution Pool, (b) Cognitive Reasoning Layer, and (c) Interactive Feedback-driven Layer.

\subsubsection{\textbf{Instruction Evolution Pool (IEP)}}
\leavevmode\par\noindent
The IEP is used to increase the complexity of TS multimodal data, aiming to activate latent neural units in LLMs related to TS tasks. Based on~\cite{ITFormer,Chatts,NNLCB,PMFFRC}, we summarize the following 7 primary evolution attribute strategies and 55 secondary evolution directions for TS instruction data.

\textbf{\ding{172} Trend}: Evolving the variation trends of historical time-series data, including upward, downward, stable, and mixed.

\textbf{\ding{173} Seasonality}: Evolving the cyclical fluctuation patterns of TS data, including none, sine wave, square wave, triangular wave, sawtooth wave, stepwise, composite, random, and segmented.

\textbf{\ding{174} Statistics}: Quantifying the numerical characteristics of TS data, consisting of mean, variance, standard deviation, maximum, minimum, skewness, unit root test value, and mode.

\textbf{\ding{175} Local Features}: Identifying the key points or anomalous behaviors, including local peaks, troughs, abrupt changes, outliers, spikes, jumps.

\textbf{\ding{176} Multivariate}: Including covariance matrix, granger causality, mutual information, earth mover’s distance, TS alignment distance et al.

\textbf{\ding{177} Compressed Pepresentation}: Generating compact representations of raw data by using discrete wavelet transform huffman encoding et al.

\textbf{\ding{178} Background Evolution}: Expanding and reinforcing non-time-series content through synonym substitution, background rewriting, background elaboration, and keyword condensation.

\subsubsection{\textbf{Cognitive Reasoning Layer (CRL)}}

\leavevmode\par\noindent
As shown in Fig.~\ref{tsevol}(b), CRL aims to enhance the reasoning  of subsequent MTSLLM by generating high-quality time–text data. Due to the high cost of constructing such data, it leverages a reasoning LLM like DeepSeek-R1~\cite{guo2025deepseek} and ChatGPT-o3~\cite{lin2025performance} to create the enhanced TS instruction data, which is then further learned by the target MTSLLM. This essentially follows a knowledge distillation paradigm involving two dialogue agents, Consultant and Client.

\textbf{Consultant Agent}: This agent generates guiding questions via the seven evolution attributes defined in the IEP, steering the trajectory of the multi-turn dialogue. The CRL's design is inspired by psychological counseling paradigms~\cite{zhang2025counselor}, where the Consultant leverages a frontier LLM to emulate a therapist’s cognitive guidance.

\textbf{Client Agent}: Serving as the counterpart to the Consultant, the Client is responsible for generating precise responses to the elicited questions. Leveraging the advanced reasoning capabilities of models like ChatGPT-5.1, the Client’s outputs are encapsulated within <\textit{think}>...</\textit{think}> tags to explicitly manifest its Chain-of-Thought (CoT). This transparent reasoning process significantly bolsters the subsequent inference and multi-turn dialogue capabilities of the MTSLLM. Details of CoT-driven interaction is provided in Table~\ref{TSEvol-DEMO}, {\color{black}Appendix~\ref{ExamplesofQuestionAnswerPairs}}.

\subsubsection{\textbf{Interactive Feedback-driven Layer }}
\leavevmode\par\noindent
The IFL mechanism enhances the quality of CRL data generation via a dialogue framework between Supervisor and Intern. Its design resembles supervised filtering, but instead of relying on an explicit reward or penalty function~\cite{liu2024deepseek}, feedback is automatically inferred by role-play agent.

\textbf{Intern Agent}:
This agent emulates the learning process of an observer acquiring knowledge from the therapeutic dialogues between the Client and Consultant. Ideally, the Intern’s architecture is aligned with the target MTSLLM, a design choice intended to pre-activate domain-specific data pathways and TS-related latent neurons, thereby ensuring seamless knowledge transfer during the subsequent fine-tuning phase.

\textbf{Supervisor Agent}: The Supervisor evaluates the quality of each evolution loop (a complete Q\&A pair) and provides implicit feedback to guide subsequent question generation. As defined in Eq.~(\ref{eq:ad}), the evaluation $G_k$ is a binary alignment check:
\begin{equation}
\label{eq:ad}
\begin{split}
G_k = \left\{
\begin{array}{l}
True\;; \;\;if \; D_k \Leftrightarrow T,\\
False\;; \text{\;\;otherwise}.
\end{array}
\right.
\end{split}
\end{equation}
Here, $\Leftrightarrow$ is the alignment between the Intern's answer $D_k$ in the $k$-th round and the ground truth $T$, across $\rho$ (default 3) user-defined evolution rounds. The complete logic and prompts for TSEvol's four-agent interaction are detailed in {Appendix~\ref{Details of Algorithm} Algorithm~\ref{TSEvol-ALGORITHM}} and {Appendix~\ref{ExamplesofQuestionAnswerPairs} Tables~\ref{ConsultantPromopt}-\ref{InternPromopt}}.

\subsection{ChatAD Model Training}
As shown in Fig.~\ref{mtsllm}(b), we propose a two-stage training paradigm for ChatAD: initial SFT for AD task using TSEData, followed by TKTO optimization to bolster reasoning and generalization.

\subsubsection{SFT for Anomaly Detection Task}
We adopt LoRA (Low‑Rank Adaptation)~\cite{hu2022lora}-based SFT to train the adapter. In this stage, we contribute three models using TSED-20K as shown in Table~\ref{parameters}, including ChatAD-Llama3-8B, $\sim$Qwen2.5-7B, and $\sim$Mistral-7B. The choice of base models was motivated by the following considerations: 1) they are commonly used in both industry and academia, and medium-sized models offer lower resource consumption for training and inference; 2) alignment with benchmark methods like TimeMQA; 3) explore the applicability of TSEvol across different models. 

{
\begin{table}[t!]
\tabcolsep=0.08cm       % liejianju
\centering
\caption{Hyper-parameter details of the ChatAD family}
\resizebox{\linewidth}{!}{
\begin{tabular}{llll}
\hline
Hyperparameter & \multicolumn{3}{c}{ChatAD} \\ \hline
Base model     & $\sim$Llama3-8B & $\sim$Qwen2.5-7B & $\sim$Mistral-7B\\
Learning rate & 5e-5 & 5e-5 & 5e-5 \\
Lr scheduler & cosine & cosine & cosine\\
Batch size & 4 & 8 & 4\\
Training epochs & 5& 5 & 5 \\
Lora rank & 16 & 16 & 16 \\
Lora alpha & 16 & 16 & 16 \\
%lora dropout & 0 & 0 & 0\\
Max cutoff length & 8192 & 8192 & 8192\\ 
Cptimizer & \multicolumn{3}{l}{Adamw\_touch, betas=(0.9, 0.999)} \\
Target modules & \multicolumn{3}{l}{gate, up, down, v, q, o, k }\\
Training device & \multicolumn{3}{l}{NVIDIA A100 (80 GB) GPUs}\\
Training time & $\sim$41 hours & $\sim$36 hours & $\sim$62 hours \\
\hline
\end{tabular}
}
\label{parameters}
\end{table}
}

\subsubsection{TKTO for Enhanced Generalization}
We propose TKTO, which leverages TS features to boost ChatAD's cross-task generalization performance. 
Let $(x, y)$ denote the Q\&A pairs. To quantify the quality of each pair, we define a preference vector $d = [d_0, d_1, d_2, d_3]$, where 
$d_0 \in [0, 25]$ is the LLM-Feedback (LLMF) score for format compliance; 
$d_1 \in [0, 50]$ is the LLMF score for TS task correctness; 
$d_2 \in [0, 25]$ is the LLMF score for answer fluency; and 
$d_3 \in [0, 100]$ is the Human Preference Feedback (HPF) score. 
Detailed scoring criteria are provided in Appendix~\ref{TKTO-SUP} Table~\ref{LLM-TKTO}.
We first calculate a composite score $S$ as the weighted sum of these feedbacks:
\begin{equation}
\label{eq:score_s}
S = \epsilon \cdot \underbrace{(d_0+d_1+d_2)}_{\text{LLMF}} + (1-\epsilon) \cdot \underbrace{d_3}_{\text{HPF}},
\end{equation}
where $\epsilon$ (default 0.5) is the weighting factor. To handle boundary noise inherent in hard-threshold labeling, we transform $S$ into a continuous preference score $\alpha \in [0, 1]$ using a sigmoid mapping ($\sigma$) as TS-related probabilistic soft embedding:
\begin{equation}
\label{eq:alpha_mapping}
\alpha = \sigma\left( \kappa \cdot (S - \eta) \right).
\end{equation}
Here, $\eta$ (default 60) is the threshold center, and $\kappa > 0$ is a fixed steepness coefficient (default 0.1) that controls the softness of the transition. 
A value of $\alpha$ closer to 1 indicates the pair $(x, y)$ is strongly preferred, while $\alpha$ closer to 0 indicates a rejected or poor-quality sample.

Finally, the TKTO loss function is defined as:
{\small
\begin{equation}
\label{eq:tkto}
\begin{split}
\mathcal{L}_{\text{TKTO}}(\pi_{\theta}) = \mathbb{E}_{(x,y,d)\sim \mathcal{D}}[\alpha \cdot \mathcal{L}_D(\pi_{\theta}, x, y) + \\(1-\alpha) \cdot \mathcal{L}_U(\pi_{\theta}, x, y)],
\end{split}
\end{equation}}

where $\mathcal{L}_D(\pi_{\theta}, x, y) = -\log \sigma ( \beta \cdot \tau(x, y) )$ and $\mathcal{L}_U(\pi_{\theta}, x, y) = -\log \sigma ( -\beta \cdot \tau(x, y) )$ are desirable (preference) and undesirable (rejected) losses in standard KTO algorithm. 
The terms $\sigma$, $\beta$, and $\tau(x, y) = \log \frac{\pi_{\theta}(y|x)}{\pi_{\text{ref}}(y|x)}$ denote the sigmoid function, the regularization coefficient, and the log probability ratio of the current model $\pi_{\theta}$ (ChatAD*) relative to the reference model $\pi_{\text{ref}}$ (ChatAD).

\subsection{LLM-driven LAD Benchmark}

\subsubsection{\textbf{Datasets}}\leavevmode\par\noindent

{
\small
\begin{table}[!]
\tabcolsep=0.08cm       % liejianju
\centering
\caption{Details of Datasets. ``$^{\dagger}$'' denote contributed by this study. Agents for TSEData are initialized with Gpt-5 family, consuming about 234.05M tokens ($\sim$2K\$).}
\resizebox{\linewidth}{!}{
\begin{tabular}{lcc}
\hline
Name & Number & Short Description \\ 
\hline
SGAD$^{\dagger}$ & 16044 & smart grid dataset for generalization task\\
TSEData$^{\dagger}$ & 21303 & $\sim$90\% used to train ChatAD, $\sim$10\% used \\
 &  &  to evaluate multi-turn dialogue capability \\
TKTOD$^{\dagger}$ & 9976 & RL dataset with LLM and human feedback, \\
& & used to build a generalized ChatAD*. \\
\hdashline
ANDE & 37000 & $\sim$80\% used to train ChatAD, $\sim20$\% used
to \\
&&evaluate single-turn AD effectiveness \\
ClASS & 37000 & classification dataset for generalization task \\
OERQA & 37629 & open-ended QA for reasoning ability task \\
IMPUT & 38657 & imputation dataset for generalization task \\
FOREC & 42557 & forecasting dataset for generalization task \\
\hline
\end{tabular}
}
\label{datasets}

\end{table}
}
{\small
\begin{table*}[t!]
\tabcolsep=0.08cm
\centering
\caption{Results of single-turn AD task on ANDE dataset and multi-turn dialogue task on TSEData. The $^{\#\{1-5\}}$ and boldface denote the Top-5 and the best result except for ChatAD* (the ChatAD-Qwen2.5-7B with TKTO).}
{
\resizebox{\linewidth}{!}{
\begin{tabular}{lrrr|rrr}
\hline
\multirow{2}{*}{Models / Metrics}& \multicolumn{3}{c}{The Anomaly Detection Effectiveness Task} &\multicolumn{3}{|c}{The Multi-turn Dialogue Capability for AD Task}\\ \cline{2-7}
 & Accuracy $\uparrow$ (\%) & F1 Score $\uparrow$ (\%) & FPR $\downarrow$ (\%) & Accuracy$\uparrow$ (\%) & F1 Score $\uparrow$ (\%) & FPR $\downarrow$ (\%)\\ \hline
Mistral-7B-Instruct & 51.53 & 60.00 & 69.47 & 70.17 & 75.70 & 53.03\\
Llama3-8B-Instruct & 51.42 & 59.80 & 68.97 & 69.63 & 73.88 & 48.66\\
Qwen2.5-7B-Instruct & 54.35 & 52.39 & 40.66 & 69.42 & 73.31 & 47.32\\
Qwen3-8B & 53.30 & 39.23 & 22.41 & 60.86 & 70.66 & 69.98\\
DeepSeek-R1-Distill-Llama3-8B & 51.50 & 61.86 & 75.32 & 60.18 & 68.12 & 66.25\\
DeepSeek-R1-Distill-Qwen2.5-7B & 55.28 & 57.12 & 48.41 & 50.99 & 65.87 & 92.77 \\
TimeMQA-Llama3-8B & $^{\#{4}}$ 70.42 & $^{\#{4}}$ 67.15 & $^{\#{4}}$ 18.59 & $^{\#{4}}$ 75.77 & $^{\#{4}}$ 77.38 & $^{\#{4}}$ 33.74 \\
TimeMQA-Mistral-7B & 61.77 & 55.94 & 24.48 & 66.11 & 67.11 & 40.47 \\
TimeMQA-Qwen2.5-7B & $^{\#{5}}$ 69.97 & $^{\#{5}}$ 62.80 & $^{\#{3}}$ 9.41 & $^{\#{5}}$ 70.30 & $^{\#{5}}$ 71.74 & $^{\#{5}}$ 37.32 \\
(Ours) ChatAD-Llama3-8B & $^{\#{3}}$ 75.79 & $^{\#{3}}$ 74.82 & $^{\#{5}}$ 19.17 & $^{\#{3}}$ 82.38 & $^{\#{3}}$ 83.96 & $^{\#{3}}$ 27.70 \\
(Ours) ChatAD-Mistral-7B & $^{\#{2}}$ 90.54 & $^{\#{2}}$ 90.31 & $^{\#{2}}$ 5.79 & $^{\#{1}}$ \textbf{96.46} & $^{\#{1}}$ \textbf{96.32} & $^{\#{2}}$ 3.90 \\
(Ours) ChatAD-Qwen2.5-7B & $^{\#{1}}$ \textbf{92.23} & $^{\#{1}}$ \textbf{92.08} & $^{\#{1}}$ \textbf{4.55} & $^{\#{2}}$ 95.69 & $^{\#{2}}$ 95.39 & $^{\#{1}}$ \textbf{1.95}\\ 
\hdashline
ChatAD* ($\sim$Qwen2.5-7B + TKTO) & 92.23 & 92.03 & 5.18 & 96.07 & 95.83 & 2.19 \\
\hline
\end{tabular}
}}
\label{ADTASK}
\end{table*}}
Table~\ref{datasets} summarizes experimental datasets. TSEData is an AD dataset evolved from ANDE through TSEvol, integrating multiple public series (e.g., NASA, UCR, Yahoo)~\cite{han2022adbench,zamanzadeh2024deep}. SGAD is a novel smart grid dataset designed to prevent LLM contamination. Regarding Q\&A and reasoning, CLASS provides multi-domain dialogue data for TSQA~\cite{timemqa}, while OERQA (from Time-MQA) offers open-domain reasoning across natural and internet sources. IMPUT and FOREC complete the suite, focusing on healthcare and human activity domains. Additionally, TKTOD serves as a RL dataset based on human feedback (RLHF) for TKTO. See {Appendix~\ref{Details of Datasets}} for further details.

\subsubsection{\textbf{Metrics}}\leavevmode\par\noindent
As illustrated in Fig.~\ref{mtsllm}(c), LLADBench consists of four tasks. To evaluate single-turn and multi-turn dialogue capability, we employ Accuracy (Acc), F1 score, and False Positive Rate (FPR). Using standard notation for True Positives ($TP$), True Negatives ($TN$), False Positives ($FP$), and False Negatives ($FN$), these metrics are defined as~\cite{timemqa,CGPU-F3SR}:

{\small
\begin{equation}
\label{eq:accuracy}
\begin{split}
Acc\;(\%, \uparrow) = \frac{TP+TN}{TP+TN+FP+FN}\times 100\%,
\end{split}
\end{equation}}

{\small
\begin{equation}
\label{eq:f1-}
\begin{split}
F1\;(\%, \uparrow) = 2\times \frac{Precision\times Recall}{Precision+Recall}\times 100\% .
\end{split}
\end{equation}}

{\small
\begin{equation}
\label{eq:fpr}
\begin{split}
FPR\;(\%, \downarrow) = \frac{FP}{FP+TN}\times 100\%,
\end{split}
\end{equation}}

Where Recall and Precision follow standard definitions. For TS reasoning task, we utilize LLM-based scoring (LS, $\uparrow$) and format-matching scores (FS)~\cite{luo2025timera}. The adoption of an LLM-as-a-judge is necessitated by the open-ended nature of OERQA, which includes trend analysis, extrema computation, and multiple-choice et al., making traditional metrics insufficient. FS is defined as:

{
\small
\begin{equation}
\label{eq:formate}
\begin{split}
FS\;(\%, \uparrow) = \frac{Number\_parsed\_samples}{Number\_total\_samples}\times 100\% .
\end{split}
\end{equation}}

For generalization across imputation and forecasting, we use FS as shown in Eq.~(\ref{eq:formate}) and Mean Absolute Error (MAE). For multi-class parsing tasks involving complex text, we adopted LLM-based scoring as shown in {Appendix~\ref{DetailsofLLADBench} Table~\ref{RAPromopt}}.

\subsubsection{\textbf{Baseline Models}} \leavevmode\par\noindent
We benchmarked ChatAD against 9 medium-sized open-sourced models: Llama3-8B-Instruct~\cite{meta_Llama3_8b_instruct}, Qwen3-8B~\cite{qwen3technicalreport}, Mistral-7B-Instruct~\cite{jiang2024mistral}, Qwen2.5-7B-Instruct~\cite{qwen25technicalreport}, and TimeMQA~\cite{timemqa} family (TimeMQA-Llama3-8B, TimeMQA-Mistral-7B, DeepSeek-R1-Distill-Qwen2.5-7B, DeepSeek-R1-Distill-Llama3-8B~\cite{deepseek2025r1}, and TimeMQA-Qwen2.5-7B). Details can be found in {Appendix~\ref{Details of Baselines}}.

{\small
\begin{table*}[!t]
    \centering
    \caption{ TS reasoning and generalization results. SGAD, FORCE, IMPUT, and CLASS are used for cross-data/task tests (AD, forecasting, imputation, classification). DS: DeepSeek-R1-Distill. "$---$": Parsing rate < 10\%.}
    \resizebox{\linewidth}{!}{\begin{tabular}{lrr|rr|rr|rr|r}
    \hline
        %Model/Task & Cross Data & ~ & ~ & Cross Tasks Generalization Tests & ~ & ~ & ~ &   & ~ \\ \hline
        %~ & Anomaly Detection & ~ & ~ & Forcasting & ~ & Imputation & ~ & Classification  & ~ \\ \hline
        \multirow{3}{*}{Models / Tasks} & \multicolumn{2}{c|}{Reasoning Task} & \multicolumn{7}{c}{Cross-Data \& Tasks Generalization Ability Task } \\ \cline{2-10} 
        ~ & \multicolumn{2}{c|}{OERQA} & \multicolumn{2}{c|}{SGAD} & \multicolumn{2}{c|}{FOREC} & \multicolumn{2}{c|}{IMPUT} & CLASS \\ \cline{2-10} 
        ~& LS $\uparrow$ (\%) & FS $\uparrow$ (\%) & F1 $\uparrow$ (\%) & FS $\uparrow$ (\%) & MAE $\downarrow$ & FS $\uparrow$ (\%) & MAE $\downarrow$ & FS $\uparrow$ (\%) & LS$\uparrow$  \\ \hline
Mistral-7B-Instruct               & 38.92           & 95.30           & 62.45          & 89.03          & $---$          & 7.20            & $---$         & 9.43           & 32.70 \\
Llama3-8B-Instruct                & 49.11           & 99.87           & 67.11          & 99.91          & 48.52          & 49.40           & 4.23          & 17.34          & 44.30 \\
Qwen2.5-7B-Instruct               & 46.49           & $^{\#4}$ 99.96  & 64.16          & 53.66          & $^{\#2}$ 31.56 & 42.30           & 2.69          & $^{\#5}$ 74.93 & 38.10 \\
Qwen3-8B                          & 43.32           & $^{\#5}$ 99.88  & 35.71          & $^{\#5}$ 99.92 & $---$          & 0.30            & $---$         & 3.3            & 37.30 \\
DS-Llama3-8B                      & 47.18           & 97.87           & $^{\#5}$ 67.17 & 99.90          & $---$          & 3.80            & 3.87          & 28.00          & 38.50 \\
DS-Qwen2.5-7B                     & 49.93           & 95.81           & 42.83          & 77.37          & $---$          & 0.50            & 3.23          & 10.2           & 40.00 \\
TimeMQA-Mistral-7B                & $^{\#3}$ 64.08  & 99.49           & 66.62          & $^{\#3}$ 99.97 & $^{\#4}$ 41.98 & 71.40           & $^{\#4}$ 1.93 & $^{\#3}$ 80.76 & $^{\#2}$ 65.00 \\
TimeMQA-Llama3-8B                 & 55.57           & 91.75           & 67.15          & 99.90          & 42.41          & $^{\#1}$ 86.40  & $^{\#1}$ \textbf{1.14} & 71.24          & 42.70 \\
TimeMQA-Qwen2.5-7B                & $^{\#2}$ 66.03  & $^{\#3}$ 99.97  & 61.67          & $^{\#2}$ 99.98 & 42.79          & $^{\#3}$ 79.40  & $^{\#2}$ 1.21 & $^{\#2}$ 84.65 & 52.10 \\ \hdashline
(Ours) ChatAD-LLama3-8B           & $^{\#5}$ 55.64  & 95.82           & $^{\#2}$ 67.20 & $^{\#1}$ \textbf{99.99} & $^{\#5}$ 42.32 & $^{\#2}$ 84.90  & $^{\#5}$ 2.60 & $^{\#4}$ 77.56 & $^{\#3}$ 63.30 \\
(Ours) ChatAD-Mistral-7B          & $^{\#4}$ 63.31  & $^{\#1}$ \textbf{99.99}  & $^{\#3}$ 67.19 & $^{\#1}$ \textbf{99.99} & 50.15          & $^{\#4}$ 77.80  & 2.94          & 40.25          & $^{\#5}$ 44.10 \\
(Ours) ChatAD-Qwen2.5-7B          & 55.45           & $^{\#2}$ 99.98  & $^{\#4}$ 67.18 & $^{\#3}$ 99.97 & $^{\#3}$ 41.48 & 41.50           & 3.30          & 65.65          & $^{\#4}$ 55.00 \\
ChatAD* ($\sim$Qwen2.5 + TKTO)    & $^{\#1}$ \textbf{69.83}  & $^{\#1}$ \textbf{99.99}  & $^{\#1}$ \textbf{68.13} & $^{\#4}$ 99.96 & $^{\#1}$ \textbf{29.04} & $^{\#5}$ 74.60  & $^{\#3}$ 1.84 & $^{\#1}$ \textbf{88.58} & $^{\#1}$ \textbf{81.90} \\
        \hline
    \end{tabular}}
    \label{reasoning-generalization}
\end{table*}}

\section{Experimental Results}\label{Benchmark Results}
%This section first presents the LLADBench results, followed by the ablation and additional results.
\subsection{LLADBench Results}
\subsubsection{\textbf{Anomaly Detection Effectiveness}} 
Table~\ref{ADTASK} (left) demonstrates that ChatAD-Qwen2.5-7B outperforms all modles. Specifically, it achieves substantial improvements in Accuracy (21.81\% $\sim$ 40.81\%) and F1 score (24.93\% $\sim$ 52.85\%) over competitors ranging from TimeMQA-Llama3-8B to Llama3-8B-Instruct, while simultaneously reducing the FPR by 4.86\% to 70.77\%. Other family members, ChatAD-Mistral-7B and ChatAD-Llama3-8B, also exhibit superior performance, yielding average gains of 18.06\% in Acc and 32.88\% in F1 with significant FPR reductions.

The superiority of the ChatAD family in single-turn AD dialogues is twofold: 1) extensive instruction tuning on approximately 50K multi-scenario dialogue turns enhances general AD proficiency, and 2) the integration of TSEvol-evolved data significantly bolsters the model's domain-specific comprehension of complex anomaly scenarios.

\subsubsection{\textbf{Multi-turn Dialogue Capability for AD}}\leavevmode\par\noindent
Table~\ref{ADTASK} (right) reveals that multi-turn dialogue consistently boosts AD performance across all models. For instance, Mistral-7B-Instruct experiences a substantial performance leap when transitioning from single-turn to multi-turn dialogue, with Accuracy and F1 scores rising from 51.53\% and 60.00\% to 70.17\% (+18.67\%) and 75.70\% (+15.70\%), respectively, while the FPR decreases by 16.44\%.

The ChatAD family continues to dominate the multi-turn setting. Its three variants achieve average improvements of up to 30.52\% in Acc and 24.79\% in F1, alongside an FPR reduction of up to 52.44\%. These results underscore the significant potential of TSEvol: by enriching multi-turn contextual dialogues, it provides a robust framework for aligning LLMs with intricate AD requirements.

\subsubsection{\textbf{Time Series Reasoning Ability}}\leavevmode\par\noindent
The results in Table~\ref{reasoning-generalization} (left) show that ChatAD* (ChatAD-Qwen2.5-7B + TKTO) achieves the best LS and FS, reaching 69.83 and 99.99\%, respectively. In LS, ChatAD* outperforms the 2nd to 5th ranked models by a significant margin of 3.80\% (TimeMQA-Qwen2.5-7B) to 14.19\% (ChatAD-LLama3-8B).
This phenomenon highlights the superior performance of TKTO in evolving the original ChatAD model. Furthermore, even without the use of TKTO, the ChatAD series still achieves competitive LS scores, surpassed only by TimeMQA-Mistral-7B and TimeMQA-Qwen2.5-7B.

\subsubsection{\textbf{Time Series Generalization Ability}}\leavevmode\par\noindent
Table~\ref{reasoning-generalization} (right) presents the generalization results of ChatAD, ChatAD* (optimized via TKTO) and baselines on 1K random samples. Here, ChatAD* achieves top-ranked overall performance across four tasks; meanwhile, the ChatAD series without TKTO also demonstrates competitive advantages.

For the forecasting and imputation tasks, the numerical prediction performance consistently outperforms the benchmarks. For instance, the MAE of ChatAD* on the FOREC task is 2.52 and 12.97 lower than that of Qwen2.5-7B-Instruct (ranked 2nd) and TimeMQA-Mistral-7B (ranked 4th), respectively. Furthermore, the ChatAD series exhibits a strong presence in the multi-class CLASS task, occupying four of the top five spots.

We argue that the advantages of ChatAD are driven by: 1) the multi-agent design of TSEvol, which enables deep exploration of TS characteristics—even when evolved on AD data, it retains universal TS commonalities applicable to similar tasks; and 2) the cross-task optimization and weighted soft-ensemble provided by TKTO.

\subsection{Additional Results of ChatAD}
\subsubsection{\textbf{Ablation Studies}} \leavevmode\par\noindent
As shown in Table~\ref{ablation}, in ablation studies with 2K random samples, the combination of CRL + IFL in ChatAD-Llama-8B model improves the accuracy and F1-score by 3.53\% and 3.95\%, while reducing the FPR by 1.29\% compared to using CRL alone. This demonstrates the indispensability of the IFL, which validates raw data at a fine-grained, single-turn dialogue level via the interaction between the Supervisor and Intern agents.

As shown in Table~\ref{ablation-}, taking ChatAD-Qwen2.5-7B as base model, TKTO outperforms KTO by 0.83\%, 23.48\%, and 0.60\% in reasoning, prediction, and classification, respectively. Compared to cross-task SFT, it achieves further gains of 0.86\%, 7.73\%, and 1.00\%. This superiority is attributed to TKTO’s specialized TS modeling and the integration of probabilistic soft embeddings.
{
\small
\begin{table}[t!]
\tabcolsep=0.08cm       % liejianju
\centering
\caption{ChatAD Ablation studies. CRL: Cognitive Reasoning Layer, IFL: Interactive Feedback-driven Layer.}
\resizebox{\linewidth}{!}{
\begin{tabular}{lccc}
\hline
Combinations&Accuracy $\uparrow$ ($\%$)&F1 $\uparrow$ (\%) & FPR $\downarrow$ (\%)\\ \hline
Base 	& 51.42 (+18.18)  &59.80 (+9.49) & 68.97 (-40.63)\\
Base + CRL&66.37 (+3.53) 	&65.34 (+3.95) & 29.63 (-1.29)\\
Base + CRL + IFL&	\textbf{69.60} &		\textbf{69.29} & \textbf{28.34}\\
\hline
\end{tabular}
}
\label{ablation}

\end{table}
}

{
\small
\begin{table}[t!]
\tabcolsep=0.08cm       % liejianju
\centering
\caption{Results of ChatAD* ablation studies.}
\resizebox{\linewidth}{!}{
\begin{tabular}{lccc}
\hline
\multirow{2}{*}{Combinations}        & Reasoning       & Forcasting	           & Classification\\ \cline{2-4} 
                    & FS $\uparrow$   & MAE $\downarrow$   & Accuracy $\uparrow$ (\%)\\\hline
Base 	& 55.45 (+14.38)  & 41.48 (-12.44)         & 55.00 (+26.90)\\
Base + SFT        & 68.97 (+0.86)	  & 36.77 (-7.73)          & 80.80 (+1.00)\\
Base + KTO  & 69.00 (+0.83)	  &	52.52 (+23.48)	       & 81.30 (+0.60)\\
Base + TKTO (Used) & 69.83	          &	29.04                  & 81.90\\
\hline
\end{tabular}
}
\label{ablation-}

\end{table}
}
\subsubsection{Real-world Case Study of ChatAD} \leavevmode\par\noindent
{Fig.~\ref{chatad-demo}} gives an example of using ChatAD for anomaly detection in a traffic scenario, which demonstrates excellent format-following and analytical answering capabilities. It is worth noting that our corpus does not include descriptions such as ``Normal with a mild anomaly'', yet ChatAD emerges with near-human analytical ability.

{\small
\begin{figure}[t!]
  \centering
  \includegraphics[width=1\columnwidth]{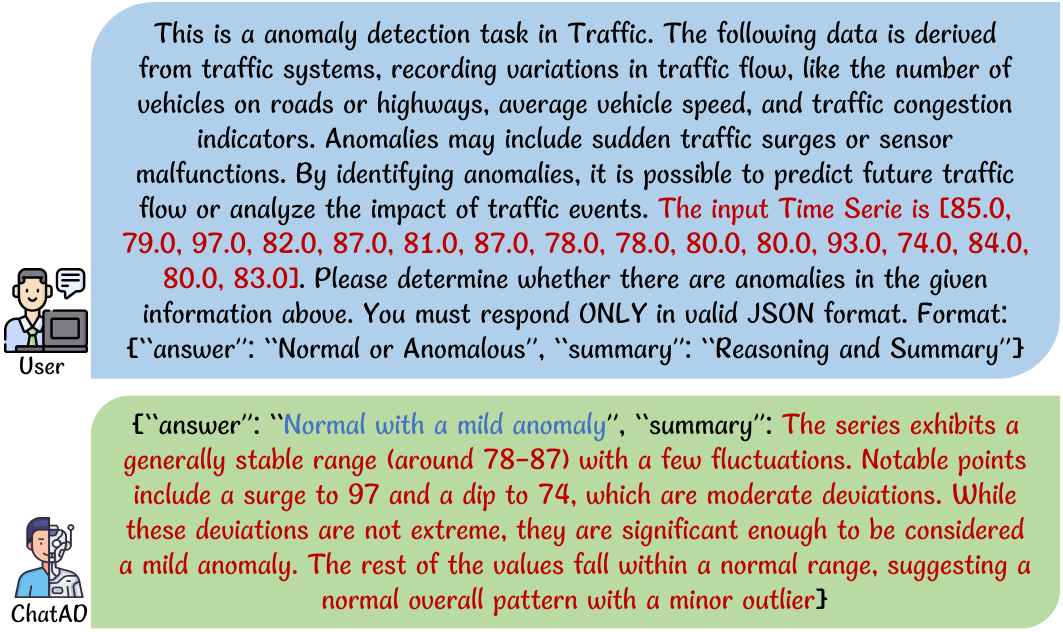}
  \caption{A real-world AD demo via using ChatAD.}
  \label{chatad-demo}
\end{figure}
}

\subsubsection{\textbf{Impact of Context Length}} \leavevmode\par\noindent
As shown in {Table~\ref{context-length}}, by progressively increasing the context length from 2048 to 8192, the ChatAD-Qwen2.5-7B's Accuracy and F1 Score continuously improved, and the FPR was significantly reduced. Since a longer context aids the model in capturing long-range dependencies, and the 8192 context length achieved the best results across all metrics, reaching an Accuracy of 92.23\%, an F1 Score of 92.08\%, and the lowest FPR of 4.55\%, thus, ChatAD adopts 8192 as the default context length configuration for the ChatAD model.

{
\begin{table}[t!]
\tabcolsep=0.08cm       % liejianju
\centering
\caption{Impact of context length of ChatAD on ANDE.}
\resizebox{\linewidth}{!}{
\begin{tabular}{lcccc}
\hline
Context Length  &Accuracy $\uparrow$ ($\%$) & F1 Score $\uparrow$ (\%)& FPR $\downarrow$ (\%)\\ \hline
ChatAD-2048	& 89.97 (+2.26) & 89.81 (+2.27) & 7.11 (-2.56)\\
ChatAD-4096 &90.97 (+1.26) 	& 90.71 (+1.37) & 4.86 (-0.31)\\
ChatAD-8192&	\textbf{92.23}  & \textbf{92.08} & \textbf{4.55}\\
\hline
\end{tabular}
}
\begin{flushleft}

\end{flushleft}
\label{context-length}

\end{table}
}

\subsubsection{\textbf{Impact of Mixed Data}}  \leavevmode\par\noindent
As shown in {\color{black}Appendix~\ref{ADDRES} Table~\ref{the-mixed-data}}, through data mixing, ChatAD series achieved more excellent performance, particularly in reducing the FPR. For example, the three series of ChatAD-Mixed reduced the FPR by 20.96\%, 45.37\%, and 49.13\% (compared to Base Model) and by 3.91\%, 1.95\%, and 1.47\% (compared to ChatAD-Unmixed), respectively. This suggests that ChatAD is more crucial in applications where minimizing the FPR is essential, as it can significantly save human resources and financial costs while maintaining system reliability.
Furthermore, the experiments also demonstrate the significant potential of the TSEvol algorithm for cross-model SFT, as its mixed data shows a certain degree of improvement for the same objective. For example, for models without TSEData mixing, ChatAD improved the F1 score by 2.32\%, 1.02\%, and 3.13\%, respectively.

\section{Conclusion and Limitations}\label{Conclusion and Future Work}
We propose TS evolution algorithm TSEvol, and introduce TSEData-20K, a large-scale multi-turn dialogue \& reasoning dataset, alongside three foundational MTSLLMs. Furthermore, we develop the TKTO algorithm to enhance the cross-task generalization of ChatAD. Finally, we establish LLADBench, where ChatAD achieves state-of-the-art performance. %For instance, ChatAD-Qwen2.5-7B demonstrates substantial improvements in accuracy ($21.81\%$--$40.81\%$ for multi-turn and $19.92\%$--$44.70\%$ for single-turn AD) and F1-score ($24.93\%$--$52.85\%$ and $19.69\%$--$29.52\%$, respectively). %Moreover, by integrating TKTO, our ChatAD exhibits a competitive advantage in cross-data/task generalization and complex TS reasoning.

ChatAD offers a novel perspective for interactive AD scenarios. However, the inherent language modeling mechanism of Transformers constrains their capacity to comprehend numerical data. Future research will focus on: 1) designing LLM architectures tailored for TS tasks to enhance time series understanding through modality alignment, and 2) integrating visual features to extend ChatAD’s capabilities in multi-modal applications.

\bibliography{custom}
\newpage
\onecolumn 
\appendix
\label{sec:appendix}
\section{Details of Algorithm}\label{Details of Algorithm}
The LLM-driven, multi-agent-based TSEvol algorithm proposed in the formal description of Algorithm~\ref{TSEvol-ALGORITHM}, where $\mathcal{A}_{c}$, $\mathcal{A}_{l}$, $\mathcal{A}_{i}$, and $\mathcal{A}_{s}$ represent the Consultant, Client, Intern, and Supervisor agents, respectively. $H$, $B$, and $T$ denote the raw time series data, text descriptions, and ground truth values, respectively. The $\rho$ represents the user-defined hyperparameter for evolution rounds, with a default configuration of 3.
{
\begin{algorithm}
\small
\tabcolsep=0.00cm       % liejianju
\setlength{\tabcolsep}{0cm}  % 更小的列间距
\renewcommand{\arraystretch}{0}  % 行高稍微压缩
\caption{\textbf{TSEvol Algorithm}}\label{TSEvol-ALGORITHM}
\KwIn{TS data $H$, short description $B$, rounds $\rho$, ground truth $T$}
\KwOut{the evolved multi-turn dialogue data $F$ with reasoning flags}
$\mathcal{A}_{c}, \mathcal{A}_{l},\mathcal{A}_{i},\mathcal{A}_{s}$ $\leftarrow$ Initialize the  \underline{c}onsultant agent, c\underline{l}ient agent, \underline{i}ntern agent, and \underline{s}upervisor agent via preset LLMs\;
$Q_0$ $\leftarrow$ $\mathcal{A}_c$ Initialize the first question from template\;
\While{$\rho-1\geq0$}{  
    \qquad  $A_0$ $\leftarrow$ $\mathcal{A}_l$($Q_0, B,H$) raise the first answer\;
    \qquad  $D_0$ $\leftarrow$ $\mathcal{A}_i$($Q_0,A_0$) raise the predicted AD task answer\;
    \qquad  $G_0$ $\leftarrow$ $\mathcal{A}_s$($D_0,T$) generate feedback via Eq.(2)\;
    \If{$G_0$ is True}{Set $\rho$ $\leftarrow$ $\rho-1$, write \{$Q_0,A_0$\} to $F$ and goto step 11\; }
    \Else{Let $\rho$ $\leftarrow$ $\rho-1$ and repeat step 3 to 10 \;}
}
Generate the QA pair index $k$ $\leftarrow$ 1\;
\While{$\rho-1\geq0$}{
    \qquad $Q_k$ $\leftarrow$ $\mathcal{A}_c(Q_{k-1},A_{k-1})$ raise the following question\;
    \qquad $A_k$ $\leftarrow$ $\mathcal{A}_l(Q_{k},B)$ response to the matched answer of $Q_k$\;
    \qquad $D_k$ $\leftarrow$ $\mathcal{A}_i(Q_{k},A_k)$ get the predicted AD task answer\;
    \qquad $G_k$ $\leftarrow$ $\mathcal{A}_s$($D_k,T$) generate feedback via Eq.(2)\;
    \If{$G_k$ is True}{
        Set $\rho$ $\leftarrow$ $\rho-1$ and $k\leftarrow k+1$\; 
        Write \{$Q_k,A_k$\} to $F$ and repeat the step 12 to 20\;}
    \Else{Let $\rho$ $\leftarrow$ $\rho-1$ and repeat step 12 to 21 \;}
}
$\{Q_k,A_k\}\leftarrow$ Generate last round QA pair via template, $D_k$ and $T$\;
Write $\{Q_k,A_k\}$ to $F$ and get the final outputs $F\leftarrow \{Q_i,A_i\}_{i=0}^{k-1}$\;
\end{algorithm}}

\section{Examples of Question \& Answer Pairs}\label{ExamplesofQuestionAnswerPairs}
Table~\ref{single-turn-sample} presents the original single-turn anomaly detection dialogue data, while Table~\ref{TSEvol-DEMO} provides the two-turn dialogue data evolved via the TSEvol algorithm.
{{\begin{tcolorbox}[colback=white,colframe=black!70,title=Original single-turn question-answer sample,fonttitle=\bfseries,coltitle=white, boxrule=1pt]
\textbf{{\color{black}Question}}: This data originates from monitoring systems and devices, and it is used to evaluate system performance or detect anomalies in time series. It also assesses the health and operational performance of devices or services. The dataset includes IOPS metrics recorded during the operation of storage systems or servers. Using this data, we can determine whether anomalies have occurred within the system. The input Time Serie is {\color{blue}[25.5, 24.0, 21.0, 26.0, 21.0, 18.5, 14.5, 13.0, 18.5, 17.5, 22.5, 21.5, 7.0, 20.0, 19.5, 26.0]}. Please determine whether there are anomalies in this time series given information above. \newline
\textbf{{\color{black}Answer}}: Based on the given information, this time series includes {\color{blue}Anomaly} Point.
\end{tcolorbox}
\captionof{table}{A real-world sample of original question \& answer pair of anomaly detection task from AIOPS area in the paper~\cite{timemqa}}
\label{single-turn-sample}}

{\begin{tcolorbox}[colback=white,colframe=black!70,title=Evolved multile-turn question-answer pairs via TSEvol algorithm,fonttitle=\bfseries,coltitle=white, boxrule=1pt]
\textbf{Total Rounds}: 2

\textbf{{\color{black}Question-1}}: This data originates from monitoring systems and devices, and it is used to evaluate system performance or detect anomalies in time series. It also assesses the health and operational performance of devices or services. The dataset includes IOPS metrics recorded during the operation of storage systems or servers. Using this data, we can determine whether anomalies have occurred within the system. The input Time Serie is {\color{blue}[25.5, 24.0, 21.0, 26.0, 21.0, 18.5, 14.5, 13.0, 18.5, 17.5, 22.5, 21.5, 7.0, 20.0, 19.5, 26.0]}. For this IOPS series, do we observe a mixed trend with non-periodic behavior, and what are its mean, variance, standard deviation, min, max, and skewness? Are the peaks at indices 4 and 16 and the trough at index 13 (7.0) anomalous via z-scores? Compared to a baseline via DTW/EMD and correlated with latency via Pearson/Granger, does this drop indicate a true incident or scheduled maintenance, and would DWT/DFT or segmented averaging improve detection? \newline

\textbf{Answer-1}: 
{\color{red}<\textit{think}>} \textit{Step}-1: {\color{blue}Calculating metrics from time series}, I've got 16 values to work with: [25.5, 24.0, 21.0, 26.0, 21.0, 18.5, 14.5, 13.0, 18.5, 17.5, 22.5, 21.5, 7.0, 20.0, 19.5, 26.0]. I'll determine the mean...; \textit{Step}-2: {\color{blue}Calculating sum and mean}, Let’s compute the sum of ...; \textit{Step}-3: {\color{blue}Calculating deviations and squares}...; \textit{Step}-4: {\color{blue}Summing squared deviations}...; \textit{Step}-5: {\color{blue}Calculating skewness and z-scores}...; \textit{Step}-6: {\color{blue}Analyzing the data trend}...{\color{red}</\textit{think}>} {\color{red}<\textit{answer}>}
Trend/periodicity: Mixed fluctuations without clear periodicity (non-periodic). {\color{blue}Metrics: mean=19.75, variance=24.19, std=4.92, min=7.0, max=26.0, skewness=-0.92 (left-skew due to the deep drop). Z-scores: index 4=+1.27, index 16=+1.27 (not anomalous);} index 13 (7.0)=-2.59 (strong outlier but not a 3σ anomaly; it would trigger at 2.5σ). Incident vs maintenance (DTW/EMD, Pearson/Granger): Insufficient evidence from IOPS alone. If DTW distance to baseline is high and latency shows positive Pearson correlation increase and Granger-causal response around the drop, treat as a true incident. If DTW/EMD shows expected pattern matching maintenance windows and latency does not rise (or falls), it is likely scheduled maintenance. Recommendation: Use local/robust methods—rolling/segmented z-scores or segmented averaging—to improve detection in this short, non-periodic series. DWT is preferable to DFT here (DFT assumes periodicity); DWT or EMD can isolate the transient drop and improve sensitivity with fewer false positives.{\color{red}</\textit{answer}>}
\newline

\textbf{Question-2}: Please determine whether there are anomalies in this time series given information above. You must respond ONLY in valid JSON format. Format: \{``answer'': ``Normal'' or ``Anomalous'', ``summary'': A concise and essential analytical process. \} \newline

\textbf{Answer-2}:{\color{red}<\textit{think}>} Creating anomaly report. I'm analyzing an abrupt drop at index 13 with a change of -14.5 followed by a spike of +13 at index 14. The change point analysis indicates a significant... I'll ensure the final {\color{blue}JSON} returns ``{\color{blue}Anomalous}'' as specified.{\color{red}</\textit{think}>} {\color{red}<\textit{answer}>}\{
                ``answer'': ``{\color{blue}Anomalous}'',
                ``summary'': ``A sharp, isolated incident occurs at t=13 with a drop to 7.0 IOPS, immediately followed by a rebound at t=14. Change-point detection (PELT) flags a dominant break at t=13. In the [value, first-difference] space, Mahalanobis distances identify t=13 and t=14 as clear outliers. Distributionally, pre (t=1–12) vs post (t=13–16) shows mean 20.29→18.13 and variance 16.52→63.73; KL(post||pre) $\sim$ 0.90 nats indicates a substantial shift, primarily increased variance. The pattern aligns with a drop-then-recover incident template. Collectively, these confirm anomaly presence. \} {\color{red}</\textit{answer}>}

\end{tcolorbox}
\captionof{table}{The two-turn real Q\&A pairs evolved through the proposed TSEvol approach, featuring enhanced multi-turn dialogue and time-series reasoning capabilities.}
\label{TSEvol-DEMO}
}

\section{Details of Agent Prompts}\label{DetailsofAgentPrompts}

{\begin{tcolorbox}[colback=white,colframe=black!70,title=Consultant Agent,fonttitle=\bfseries,coltitle=white, boxrule=1pt]
\textbf{{\color{black}1) System Prompts}}: You are Consultant tasked with generating one time-series-related question based on following seven analytical dimensions:
\begin{enumerate}[itemsep=0pt, parsep=0pt]
\item {\color{blue}trend}: used to describe the variation trends in historical data, including four types: upward, downward, stable, and mixed trends.
\item {\color{blue}seasonality}: used to describe the periodic fluctuation characteristics of historical data. It includes ten types: non-periodic, sinusoidal, square wave, triangular wave, sawtooth wave, stepwise, composite, random, and segmented patterns.
\item {\color{blue}statistics}: used to quantify the numerical characteristics of historical data. It includes eight types: mean, variance, standard deviation, maximum, minimum, skewness, unit root test value, and mode.
\item {\color{blue}local features}: used to identify key points or abnormal behaviors in historical data, including peaks, troughs, mutation points, outliers, spikes, and jumps.
\item {\color{blue}multivariate Relationships}: the multivariate attribute includes 14 types: covariance matrix, Granger causality, mutual information, Dice coefficient, Pearson correlation coefficient, edit distance, Jaccard similarity, cosine similarity, Euclidean distance, Manhattan distance, DTW (Dynamic Time Warping), Mahalanobis distance, Hamming distance, KL divergence (Kullback-Leibler), Earth Mover’s Distance (EMD), and TAD (Time Series Alignment Distance).
\item {\color{blue}compressed representations}: compression is used to generate compact representations of raw data by transforming historical data into new forms that enhance complexity. It includes four methods: Discrete Wavelet Transform (DWT), Discrete Fourier Transform (DFT), segmented averaging, and segmented mode extraction.
\item {\color{blue}background evolution}: Extend and enhance the background context of non-time-series content.
\end{enumerate}
You should generate new questions from the given 'history' instruction data, and each question should be clear, relevant, and designed to guide deeper analysis or understanding of time-series data. Finish your response in English, and question must end with a question mark ``?''.
\newline

\textbf{2) $Q_0$ Prompts Generated from Template}: {\color{blue}Please conduct an detailed anomaly detection analysis on the inputted time series data and raise a question with ONLY in valid JSON format}. Format: \{``question'': ``A concise and valuable question, strictly limited to 100 words.'' , ``summary'': ``Explain why you are asking this question, strictly limited to 200 words''.\}. \newline

\textbf{{3) \color{black}$Q_k$ ($k=\{1,2,...,\rho-1\}$) Prompts from Conversations}}: {\color{blue}Please consider the previous Answer and avoid asking duplicate Question List. Question List: \{$Q_{list}$\}, Last Question Reply/Answer \{$A$\}.} Please conduct an detailed anomaly detection analysis on the inputted time series data and raise a question with ONLY in valid JSON format. Format: \{``question'': ``A concise and valuable question, strictly limited to 100 words.'', ``summary'': ``Explain why you are asking this question, strictly limited to 200 words.'' \}.\newline

\textbf{4) $Q_{\rho-1}$ Prompts Generated from Template}: Please determine whether there are anomalies in this time series given information above. {\color{blue}You must respond ONLY in valid JSON format}. Format: \{``answer'': ``Normal or Anomalous'', ``summary'': ``A concise and essential analytical process.'' \}
        
\end{tcolorbox}
\captionof{table}{Details of the used prompts for the Consultant agent $\mathcal{A}_c$}
\label{ConsultantPromopt}
}

{\begin{tcolorbox}[colback=white,colframe=black!70,title=Intern Agent,fonttitle=\bfseries,coltitle=white, boxrule=1pt]
\textbf{{\color{black}1) The System Prompt Template}}: You are a time-series expert. Your task is to answer the given question based on the given 'history' instruction data with deep reasoning if needed, but your response must remain strictly constrained. .\newline

\textbf{{2) \color{black}The $A_k$ ($k=\{0,1,2,...,\rho-1\}$) Prompt Generated from Conversations}}: {\color{blue}The original background is: ${source\_question}$.  The generated question is: ${new\_question}$.}  Please determine whether there are anomalies in this time series given information above. You must respond ONLY in valid JSON format. Format: \{``answer'': ``Normal or Anomalous'', ``summary'': ``A concise and essential analytical process, limited to no more than 150 words.''\}
     
\end{tcolorbox}
\captionof{table}{Details of the used prompts for the Intern agent $\mathcal{A}_i$}
\label{InternPromopt}}

{\begin{tcolorbox}[colback=white,colframe=black!70,title=Client Agent,fonttitle=\bfseries,coltitle=white, boxrule=1pt]
\textbf{{\color{black}1) The System Prompt Template}}: You are a time-series expert. Your task is to answer the given question based the 'history' instruction data in a strict, accurate, and concise manner. Finish your response in English.\newline

\textbf{{2) \color{black}The $A_k$ ($k=\{0,1,2,...,\rho-1\}$) Prompt Generated from Conversations}}: {\color{blue}The original background is: ${source\_question}$.  The generated question is: ${new\_question}$.}  Please determine whether there are anomalies in this time series given information above. {\color{blue}You must respond ONLY in valid JSON format}. Format: \{``answer'': ``Provide an accurate and concise answer to the question, focusing only on the essential information, limited to no more than 200 words.\}.\newline

\textbf{3) $A_{\rho-1}$ Prompts Generated from Template}: \{``answer'': $ground\_truth$, ``summary'': ``A concise and essential analytical process from the Client agent $\mathcal{A}_{i}$'s last reply.'' \}
        
\end{tcolorbox}
\captionof{table}{Details of the used prompts for the Client agent $\mathcal{A}_l$}
\label{ClientPromopt}}

\section{Details of Datasets}~\label{Details of Datasets}
We organize all datasets in the ShareGPT format for model training and inference because it provides a better structure for multi-turn dialogue style. Figure~\ref{fig:oerqa-box} present examples of single-turn dialogue style data collected from the test set.
\begin{figure}[htbp!]
  \centering
  \begin{tcolorbox}[breakable, arc=2mm, colframe=black, boxrule=1pt,
                    left=2pt, right=2pt]
\begin{lstlisting}
{
        "conversations": [
            {
                "from": "human",
                "value": "Examine the data points [0.75, 0.74, 0.73, 0.72, 0.71, 0.69, 0.68, 0.66, 0.65, 0.63, 0.61, 0.58, 0.56, 0.54, 0.52, 0.49, 0.47, 0.45, 0.44, 0.42, 0.41, 0.4, 0.4, 0.39] and summarize the overall movement trend in this data."
            },
            {
                "from": "gpt",
                "value": "The data shows a consistent downtrend. Starting from 0.75, it decreases gradually to 0.39, indicating a steady decline over the observed period."
            }
        ]
    }
\end{lstlisting}
  \end{tcolorbox}
  \caption{Real-world example from the open-ended response question answering (OERQA) test dataset}
  \label{fig:oerqa-box}
\end{figure}

Here, we provide a detailed introduction to two new datasets introduced by LLADBench: the Smart Grid Multi-class Anomaly Detection dataset (SGAD), the General Anomaly Detection Multi-turn Dialogue dataset (TSEData), as well as the reinforcement learning dataset TKTOD for Time Series Kahneman-Tversky Optimization.

\subsection{Smart Grid Multi-class Anomaly Detection dataset (SGAD) }
In complex power systems, disturbances often occur as a mixture of multiple types. When different types of disturbances overlap, it can create significant challenges for detection and classification. Since each disturbance type has its own fundamental characteristics, these characteristics can be leveraged to identify the corresponding disturbance. Therefore, before classifying disturbances, it is necessary to thoroughly understand and analyze the features of different disturbance types. Based on the IEEE standard~\cite{quality1995ieee} and the methods described in~\cite{goh2022integrated}, we construct a dataset comprising five typical transient disturbances, two typical steady-state disturbances, and nine composite disturbances formed by their combinations. A detailed introduction is provided regarding the definitions of these disturbances and their descriptive characteristics.
\begin{itemize}
    \item \textbf{C0: Normal Signal}. The ideal electrical waveform should be a perfectly symmetrical sine wave, with voltage, current, and frequency all maintained within their rated ranges, without any distortion or deviation. This represents the ideal state of normal operation in a power system.
    \item \textbf{C1: Voltage Interruption}. Voltage interruption is usually associated with short-circuit faults in power systems. Sometimes, starting or switching large power loads can also cause voltage interruptions. A voltage interruption refers to a short-term amplitude disturbance during which the voltage magnitude drops below 0.1 pu, and the disturbance duration does not exceed one minute. Typically, the duration of the disturbance and the magnitude of the drop are used as the basis for its identification.
    \item \textbf{C2: Voltage Sag}. Voltage sag is usually associated with short-circuit faults in the system, but it can also be caused by switching heavy loads or starting large motors. It is one of the most common power quality disturbances. Voltage sag refers to a short-term voltage amplitude disturbance where the voltage magnitude drops to between 0.1 pu and 0.9 pu for a duration ranging from 0.5 cycle to less than one minute. Typically, the disturbance duration and the magnitude of the voltage drop are used as descriptive features.
    \item \textbf{C3: Voltage Swell}. Similar to voltage sag, voltage swell mainly occurs due to short-circuit faults. A voltage swell refers to a short-term amplitude disturbance where the voltage magnitude suddenly rises to between 1.1 pu and 1.8 pu within 0.5 cycle to one minute. Typically, the disturbance duration and the magnitude of the rise are used as its descriptive features.
    \item \textbf{C4: Oscillatory Transient}. Oscillatory transient is mainly caused by grounding faults or instantaneous switching of loads and breakers. It represents a sudden non-power-frequency change occurring under steady-state conditions of voltage or current (or both), and its polarity alternates in both directions (positive and negative). Typically, high frequency, oscillation amplitude, and disturbance duration are used as descriptive features.
    \item \textbf{C5: Impulse Transient}. Impulse transients are mainly caused by lightning. They represent a sudden non-power-frequency change under steady-state conditions of voltage or current (or both). The polarity of the sudden change is unidirectional (primarily positive or negative). The duration of impulse transients is extremely short. Based on disturbance duration, impulse transients can be categorized into three levels: nanosecond-level, microsecond-level, and millisecond-level impulse transients. Therefore, disturbance duration is the primary descriptive feature for impulse transients.
    \item \textbf{C6: Voltage Flicker}. Arc furnaces are one of the common causes of voltage flicker in transmission and distribution systems. The amplitude variation range of voltage flicker is between 0.95 pu and 1.05 pu, representing a continuous voltage fluctuation. People can perceive this fluctuation through changes in lamp brightness. Voltage flicker is usually characterized by the amplitude of voltage fluctuation and the number of fluctuations. The number of fluctuations relates to the count of peaks and troughs in the flicker curve; each pair of adjacent maximum and minimum values constitutes one voltage fluctuation.
    \item \textbf{C7: Harmonic}. In power systems, the nonlinear characteristics of power electronic devices cause harmonics. A harmonic is a sinusoidal component whose frequency is an integer multiple of the rated frequency. When harmonic components combine with the fundamental frequency component, waveform distortion occurs. With the increasing application of power electronic devices, harmonic distortion has become more severe, negatively impacting power users and the entire power system, such as reducing equipment accuracy and causing instability in system operation.
    \item \textbf{C8: Voltage Sag with Harmonics}. A composite disturbance phenomenon where voltage sag and harmonics occur simultaneously. Voltage sag causes a reduction in RMS voltage, while harmonics distort the waveform. The combination of these two effects has a greater impact on sensitive equipment. Common scenarios include areas with dense power electronic devices, such as semiconductor manufacturing plants and data centers.
    \item \textbf{C9: Voltage Sag with Harmonics}. A composite disturbance phenomenon where voltage sag and harmonics occur simultaneously. Voltage sag causes a reduction in RMS voltage, while harmonics distort the waveform. The combination of these two effects has a greater impact on sensitive equipment. Common scenarios include areas with dense power electronic devices, such as semiconductor manufacturing plants and data centers.
    \item \textbf{C10: Voltage Interruption with Harmonics}. A composite disturbance where voltage interruption and harmonics occur simultaneously. Voltage interruption leads to complete or nearly complete power supply loss, while harmonics cause waveform distortion, severely affecting precision electronic equipment. For example, an uninterruptible power supply (UPS) failure in a data center can result in power interruption, and harmonics may damage the UPS and connected electrical devices.
    \item \textbf{C11: Voltage Flicker with Harmonics}. Composite disturbance where flicker and harmonics occur simultaneously. Flicker causes periodic changes in lighting brightness, while harmonics distort the waveform, jointly affecting visual comfort and equipment operation. Common in environments where impact loads such as arc furnaces and rolling mills work together with power electronic devices.
    \item \textbf{C12: Harmonics with Oscillatory Transient}. Composite disturbance where harmonics and oscillatory transients occur simultaneously. Harmonics cause waveform distortion, while oscillatory transients introduce system resonance, which may lead to harmonic amplification.
    \item \textbf{C13: Voltage Flicker with Oscillatory Transient}. Composite disturbance where flicker and oscillatory transient occur simultaneously. Flicker causes changes in lighting brightness, while oscillatory transient introduces system resonance. Together, they impact overall system stability.
    \item \textbf{C14: Impulse with Harmonics}. Composite disturbance where impulse and harmonics occur simultaneously. Impulse causes instantaneous voltage spikes, while harmonics distort the waveform, potentially damaging insulation layers of power electronic equipment.
    \item \textbf{C15: Voltage Sag with Oscillatory Transient}. Composite disturbance phenomenon where voltage sag and oscillatory transient occur simultaneously. Voltage sag leads to a reduction in RMS voltage, while the oscillatory transient introduces system resonance, potentially amplifying the disturbance impact.
    \item \textbf{C16: Voltage Swell with Oscillatory Transient}. Composite disturbance where voltage swell and oscillatory transient occur simultaneously. Voltage swell increases RMS voltage, while oscillatory transient introduces system resonance, potentially accelerating equipment damage.
\end{itemize}

\subsection{General Anomaly Detection Multi-turn Dialogue Dataset (TSEData)}
TSEData-20K comprises 21,303 high-quality multi-turn dialogue samples refined from 10,011 original entries, generated via the TSEvol algorithm utilizing the ChatGPT model series.
The TSEvol algorithm is implemented in the $TSEvol.py$ script, with the following hyperparameter configurations:
\begin{itemize}[noitemsep, topsep=1pt]
\item[]$--data\_path$: Path to the raw anomaly detection dataset to be evolved.
\item[]$--saved\_path$: The destination path where the evolved data is stored.
\item[]$--rounds$: Maximum number of evolution rounds (default: 3).
\item[]$--last\_log\_path$: The log file utilized for resuming from breakpoints.
\item[]$--consultant\_model$: Model for the Consultant agent in TSEvol.
\item[]$--client\_model$: Model for the Client agent in TSEvol.
\item[]$--intern\_model$: Model for the Intern agent in TSEvol.
\end{itemize}
A representative example is described below:
\begin{lstlisting}
# Evolve anomaly detection datasets using the TSEvol algorithm.
nohup python3 -u TSEvol.py 
  --data_path "${raw anomaly detection dataset}" 
  --saved_path "${destination path where the evolved data is stored}" 
  --rounds "${maximum number of evolution rounds}" 
  --consultant_model "gpt-5_2025-08-07" 
  --client_model "gpt-5.1_2025-11-13" 
  --intern_model "gpt-4o_2024-11-20" 
  > "${saved log path}.log" 2>&1 &
\end{lstlisting}

\subsection{Time Series Kahneman-Tversky Optimization Dataset (TKTOD)}
TKTOD contains 10,000 original cross-task time series samples, including 4,000 samples from OERQA, and 2,000 samples each from CLASS, IMPUT, and FPREC~\cite{timemqa}, as shown in Table~\ref{datasets}. 
Among them, 5,000 samples are used for TKTO ablation experiments to construct the original ChatAD* model, while the other 5,000 samples are used to build reinforcement learning data with large model and human feedback based on ChatAD*-Qwen2.5-7B after SFT using first 50000 samples. The prompts for the large language model are shown in Supplementary Table~\ref{LLM-TKTO}

\section{Details of Baselines}~\label{Details of Baselines}
This section provides details on the LLADBench baseline models and scripts.
\subsection{Mistral-7B-Instruction-v0.2}
Mistral-7B-Instruct-v0.2~\cite{jiang2024mistral} is an advanced open-source language model developed by Mistral AI (\url{https://mistral.ai/}), optimized for instruction-following and multi-turn dialogue. Built on the Mistral-7B-v0.2 architecture, it features 7.3 billion parameters and a 32,768-token context window, enabling efficient handling of long documents and complex conversations. The model employs Grouped-Query Attention (GQA) and Rotary Position Embeddings (RoPE) for improved scalability and performance, while maintaining low memory overhead.
The following script provides an example of downloading the model, performing multi-turn dialogue inference using Llama-Factory~\cite{zheng2024Llamafactory}, and analyzing the results with LLADBench.
\begin{lstlisting}
# 1) Download the mistral model
git clone https://www.modelscope.cn/AI-ModelScope/Mistral-7B-Instruct-v0.2.git
# 2) Multi-turn Dialogue Reasoning using Mistral-7B-Instruction-v0.2
Llamafactory-cli train --stage sft --do_predict --model_name_or_path ${user-defined model path} --eval_dataset ChatAD_Test --dataset_dir ${User-defined dataset path} --template mistral --output_dir ${user-defined output path} --overwrite_cache --overwrite_output_dir --cutoff_len 8192 --preprocessing_num_workers 8 --per_device_eval_batch_size 16 --predict_with_generate
# 3) Analyze Results using the LLADBench Script
python3 Multi_Turn_Digo_AD_Task_Evaluation.py --input_dir "${user-defined output path}/generated_predictions.jsonl" --output_dir "${user-defined output path}/results.json"
\end{lstlisting}

\subsection{Llama-3-8B-Instruct}
Llama-3-8B-Instruct~\cite{meta_Llama3_8b_instruct} is a state-of-the-art large language model developed by Meta (\url{https://www.meta.com}) as part of the Llama 3 family. It contains 8 billion parameters, optimized for instruction-following tasks such as question answering, summarization, and multi-turn dialogue. Built on an advanced transformer architecture, it leverages extensive pretraining on diverse datasets and fine-tuning with human feedback to improve alignment and usability. The model supports multiple languages, demonstrates strong reasoning capabilities, and is designed for efficiency, making it suitable for deployment on resource-constrained environments while maintaining high performance. Llama-3-8B-Instruct is widely used for research and production applications in natural language understanding and generation.
The following script provides an example of downloading the model, performing multi-turn dialogue inference using Llama-Factory, and analyzing the results with LLADBench.
\begin{lstlisting}
# 1) Download the Llama3 model
git clone https://www.modelscope.cn/models/LLM-Research/Meta-Llama-3-8B-Instruct
# 2) Multi-turn Dialogue Reasoning using Llama-3-8B-Instruct
Llamafactory-cli train --stage sft --do_predict --model_name_or_path ${user-defined model path} --eval_dataset ChatAD_Test --dataset_dir ${User-defined dataset path} --template Llama3 --output_dir ${user-defined output path} --overwrite_cache --overwrite_output_dir --cutoff_len 8192 --preprocessing_num_workers 8 --per_device_eval_batch_size 16 --predict_with_generate
# 3) Analyze Results using the LLADBench Script
python3 Multi_Turn_Digo_AD_Task_Evaluation.py --input_dir "${user-defined output path}/generated_predictions.jsonl" --output_dir "${user-defined output path}/results.json"
\end{lstlisting}

\subsection{Qwen2.5-7B-Instruct}
Qwen‑2.5‑7B‑Instruct~\cite{qwen2_5_7b_instruct} is an 8‑billion-parameter (7.6 B non-embedding) instruction-tuned causal language model in Alibaba Cloud’s Qwen 2.5 series (\url{https://www.alibabacloud.com/blog/601782}). It delivers significant enhancements in knowledge, coding, and mathematical reasoning, leveraging specialized expert models. The model excels in instruction following, generating long outputs (up to 8K tokens), interpreting structured data like tables, and producing structured outputs such as JSON. With a massive context window of up to 128K tokens, it handles complex and lengthy tasks efficiently. It is multilingual, supporting more than 29 languages including Chinese, English, French, Spanish, and Arabic. Licensed under Apache‑2.0, Qwen‑2.5‑7B‑Instruct balances compact model size with strong performance, making it suitable for varied applications like chatbots, structured-data processing, summarization, code generation, and math problem solving.
The following script provides an example of downloading the model, performing multi-turn dialogue inference using Llama-Factory, and analyzing the results with LLADBench.
\begin{lstlisting}
# 1) Download the Llama3 model
git clone https://www.modelscope.cn/Qwen/Qwen2.5-7B-Instruct.git
# 2) Multi-turn Dialogue Reasoning using Qwen2.5-7B-Instruct
Llamafactory-cli train --stage sft --do_predict --model_name_or_path ${user-defined model path} --eval_dataset ChatAD_Test --dataset_dir ${User-defined dataset path} --template qwen3 --output_dir ${user-defined output path} --overwrite_cache --overwrite_output_dir --cutoff_len 8192 --preprocessing_num_workers 8 --per_device_eval_batch_size 16 --predict_with_generate
# 3) Analyze Results using the LLADBench Script
python3 Multi_Turn_Digo_AD_Task_Evaluation.py --input_dir "${user-defined output path}/generated_predictions.jsonl" --output_dir "${user-defined output path}/results.json"
\end{lstlisting}

\subsection{Qwen3-8B}
Qwen‑3‑8B~\cite{qwen25technicalreport} is a cutting-edge 8.2‑billion‑parameter causal language model in Alibaba Cloud’s Qwen 3 series. It supports two modes—thinking for complex reasoning, math, and code, and non‑thinking for general dialogue—within the same model. Compared to Qwen 2.5, it significantly advances reasoning, instruction adherence, agent-style tool use, and multilingual capabilities. It natively handles up to 32,768 tokens and can scale to 131,072 with YaRN. Licensed under Apache‑2.0, Qwen‑3‑8B excels in math, programming, and creative tasks, and is well-suited for chatbots, agent workflows, code synthesis, and multi-language applications.
The following script provides an example of downloading the model, performing multi-turn dialogue inference using Llama-Factory, and analyzing the results with LLADBench. The following script provides an example of downloading the model, performing multi-turn dialogue inference using Llama-Factory, and analyzing the results with LLADBench.
\begin{lstlisting}
# 1) Download the Llama3 model
git clone https://www.modelscope.cn/Qwen/Qwen3-8B.git
# 2) Multi-turn Dialogue Reasoning using Qwen3-8B
Llamafactory-cli train --stage sft --do_predict --model_name_or_path ${user-defined model path} --eval_dataset ChatAD_Test --dataset_dir ${User-defined dataset path} --template qwen3_nothink --output_dir ${user-defined output path} --overwrite_cache --overwrite_output_dir --cutoff_len 8192 --preprocessing_num_workers 8 --per_device_eval_batch_size 16 --enable_thinking False --predict_with_generate
# 3) Analyze Results using the LLADBench Script
python3 Multi_Turn_Digo_AD_Task_Evaluation.py --input_dir "${user-defined output path}/generated_predictions.jsonl" --output_dir "${user-defined output path}/results.json"
\end{lstlisting}

\subsection{DeepSeek-R1-Distill-Llama3-8B}
DeepSeek-R1-Distill-Llama3-8B~\cite{deepseek2025r1} is an advanced distilled model derived from the DeepSeek-R1 reasoning framework and Meta’s Llama-3 architecture. With 8 billion parameters, it combines the strong language understanding of Llama-3 with DeepSeek’s reinforcement learning and reasoning optimization techniques. The distillation process transfers reasoning capabilities from larger expert models into a compact form, enabling efficient inference while maintaining high-quality performance on complex tasks such as multi-step reasoning, coding, and instruction following. Designed for cost-effective deployment, it supports long-context processing and excels in scenarios requiring logical consistency and structured outputs. This model is ideal for research and production environments where reasoning ability and resource efficiency are critical. The following script provides an example of downloading the model, performing multi-turn dialogue inference using Llama-Factory, and analyzing the results with LLADBench.
\begin{lstlisting}
# 1) Download the Llama3 model
git clone https://huggingface.co/deepseek-ai/DeepSeek-R1-Distill-Llama-8B
# 2) Multi-turn Dialogue Reasoning using DeepSeek-R1-Distill-Llama-8B
Llamafactory-cli train --stage sft --do_predict --model_name_or_path ${user-defined model path} --eval_dataset ChatAD_Test --dataset_dir ${User-defined dataset path} --template deepseekr1 --output_dir ${user-defined output path} --overwrite_cache --overwrite_output_dir --cutoff_len 8192 --preprocessing_num_workers 4 --per_device_eval_batch_size 4 --enable_thinking False --predict_with_generate
# 3) Analyze Results using the LLADBench Script
python3 Multi_Turn_Digo_AD_Task_Evaluation.py --input_dir "${user-defined output path}/generated_predictions.jsonl" --output_dir "${user-defined output path}/results.json"
\end{lstlisting}

\subsection{DeepSeek-R1-Distill-Qwen-7B}
DeepSeek-R1-Distill-Qwen-7B~\cite{deepseek2025r1} is a distilled reasoning model built on Alibaba’s Qwen architecture and enhanced by DeepSeek’s reinforcement learning and reasoning optimization techniques. With 7 billion parameters, it inherits Qwen’s strong multilingual and instruction-following capabilities while integrating advanced reasoning skills from larger expert models through distillation. This approach significantly reduces computational cost without sacrificing performance on complex tasks such as multi-step reasoning, coding, and structured data generation. The model supports long-context processing, making it suitable for applications like intelligent assistants, problem-solving agents, and research environments where efficiency and logical consistency are critical. Its design emphasizes balanced performance, scalability, and resource efficiency for real-world deployment. The following script provides an example of downloading the model, performing multi-turn dialogue inference using Llama-Factory, and analyzing the results with LLADBench.
\begin{lstlisting}
# 1) Download the Llama3 model
git clone https://huggingface.co/deepseek-ai/DeepSeek-R1-Distill-Qwen-7B
# 2) Multi-turn Dialogue Reasoning using DeepSeek-R1-Distill-Qwen-7B
Llamafactory-cli train --stage sft --do_predict --model_name_or_path ${user-defined model path} --eval_dataset ChatAD_Test --dataset_dir ${User-defined dataset path} --template deepseekr1 --output_dir ${user-defined output path} --overwrite_cache --overwrite_output_dir --cutoff_len 8192 --preprocessing_num_workers 4 --per_device_eval_batch_size 4 --enable_thinking False --predict_with_generate
# 3) Analyze Results using the LLADBench Script
python3 Multi_Turn_Digo_AD_Task_Evaluation.py --input_dir "${user-defined output path}/generated_predictions.jsonl" --output_dir "${user-defined output path}/results.json"
\end{lstlisting}

\subsection{TimeMQA Series}
Time-MQA~\cite{timemqa} is a series of multimodal question-answering models for time-series data. It introduces a unified framework that accepts natural language queries across diverse time-series tasks – ranging from classical numerical analyses (like forecasting or anomaly detection) to open-ended reasoning questions. A key innovation is the creation of TSQA, a large-scale dataset of about 200k question–answer pairs spanning 12 domains and multiple task types, which provides rich training for both numeric and explanatory QA in this domain. Each Time-MQA model is built by continually pre-training a large language model (e.g. Mistral 7B~\cite{mistral7b_instruct}, Llama-3-8B~\cite{meta_Llama3_8b_instruct}, Qwen-2.5-7B~\cite{qwen2_5_7b_instruct}) on TSQA using parameter-efficient fine-tuning (LoRA) to imbue it with specialized temporal reasoning skills. Notably, the framework incorporates context enhancement: auxiliary textual context (such as domain information) is included alongside the raw time-series input to resolve ambiguities and improve understanding. The following script provides examples of using TimeMQA for single-round anomaly detection tasks and multi-turn dialogue tasks.
\begin{lstlisting}
# 1) Download the TimeMQA Series Models
git clone https://huggingface.co/Time-MQA/Qwen-2.5-7B
git clone https://huggingface.co/Time-MQA/Mistral-7B
git clone https://huggingface.co/Time-MQA/Llama-3-8B
# 2) Multi-turn Dialogue Reasoning using TimeMQA-Mistral-7B
Llamafactory-cli train --stage sft --do_predict --model_name_or_path ${user-defined model path} --eval_dataset ChatAD_Test --dataset_dir ${User-defined dataset path} --template mistral --output_dir ${user-defined output path} --overwrite_cache --overwrite_output_dir --cutoff_len 8192 --preprocessing_num_workers 4 --per_device_eval_batch_size 4 --enable_thinking False --predict_with_generate
# 3) Analyze Results using the LLADBench Script
python3 Multi_Turn_Digo_AD_Task_Evaluation.py --input_dir "${user-defined output path}/generated_predictions.jsonl" --output_dir "${user-defined output path}/results.json"
# 4) Single-turn AD task Reasoning using TimeMQA-Qwen2.5
Llamafactory-cli train --stage sft --do_predict --model_name_or_path ${user-defined model path} --eval_dataset AD_Test --dataset_dir ${User-defined dataset path} --template mistral --output_dir ${user-defined output path} --overwrite_cache --overwrite_output_dir --cutoff_len 2048 --preprocessing_num_workers 32 --per_device_eval_batch_size 128 --predict_with_generate
# 5) Analyze AD Task Results using the LLADBench Script
python3 AD_Task_Evaluation.py --input_dir "${user-defined output path}/generated_predictions.jsonl" --output_dir "${user-defined output path}/results.json"
\end{lstlisting}

\section{Details of LLADBench}~\label{DetailsofLLADBench}
The proposed LLADBench includes a series of automated result analysis scripts. It starts from the ``*generated\_predictions.jsonl'' file produced by Llama-factory~\cite{zheng2024Llamafactory} and outputs formatted JSON. This section will provide a detailed introduction.

\subsection{Anomaly Detection Effectiveness Task Evaluation}
The $AD\_Task\_Evaluation.py$ script automates the analysis of metrics for binary anomaly detection, such as accuracy, false positive rate, and F1 score. It accepts two parameters:
\begin{itemize}[noitemsep, topsep=1pt]
\item[]$--input\_dir$, which specifies the ``*generated\_predictions.jsonl'' file produced by Llama-factory, 
\item[]$--output\_dir$, which defines the directory where the formatted JSON output will be saved.
\end{itemize}

\subsection{Multi-turn Dialogue Capability Task Evaluation}
Simlar to $AD\_Task\_Evaluation.py$, the $Multi\_Turn\_Digo\_AD\_Task\_Evaluation.py$ script accepts two parameters:
\begin{itemize}[noitemsep, topsep=1pt] 
\item[]$--input\_dir$, which specifies the ``*generated\_predictions.jsonl'' file produced by Llama-factory, 
\item[]$--output\_dir$, which defines the directory where the formatted JSON output will be saved.
\end{itemize}

\subsection{Time Series Reasoning Ability Task Evaluation}
The reasoning capability for time series consists of two components: the score for the multi-turn dialogue format and the accuracy score for open-ended question answering. The former is provided by Script $Multi\_Turn\_Digo\_AD\_Task\_Evaluation.py$, while the latter is implemented through automated parsing in Script B.

In practical testing, matching the format of answers for open-ended questions is challenging. For example, among the 1,000 randomly tested cases in this study, there were 501 open-ended questions, 303 multiple-choice questions, and 196 true/false judgments. These data span multiple scenarios, such as anomaly detection, summarization, trend analysis, volatility analysis, and seasonality investigation. To address this, we use gpt-5-2025-08-07~\cite{openai_gpt5} to automatically perform semantic analysis for score determination, with the prompt shown in Table~\ref{RAPromopt}.
{\begin{tcolorbox}[colback=white,colframe=black!70,title=Prompts of TS Reasoning Ability Task Evaluation,fonttitle=\bfseries,coltitle=white, boxrule=1pt]
You are a time-series analysis expert. Your role is to rigorously and strictly evaluate a student's answer against the given question and the ground truth.

\textbf{{\color{black}Scoring rules}}:
 \begin{itemize}[itemsep=2pt, parsep=0pt]
     \item[] - Fully correct → return 1.
     \item[] - Completely incorrect → return 0.
     \item[] - Partially correct → return a decimal between 0 and 1 (e.g., 0.75).
 \end{itemize}

\textbf{{\color{black}Output constraints}}:
 \begin{itemize}[itemsep=2pt, parsep=0pt]
     \item[] - Output only one numeric value in the format: <score>.
     \item[] - Value must be in the range [0, 1], decimals allowed.
     \item[] - Do not include explanations or any additional fields.
 \end{itemize}

\textbf{{\color{black}Inputs}}:
 \begin{itemize}[itemsep=2pt, parsep=0pt]
     \item[] - Original question: {\color{blue}$\{source\_question\}$}
     \item[] - Student answer: {\color{blue}$\{the\_predicted_answer\}$}
     \item[] - Ground truth: {\color{blue}$\{the\_ground\_truth\}$}
 \end{itemize}

\textbf{Evaluation Criteria (more lenient)}:
Focus on core time-series concepts: trend, seasonality, anomalies, volatility, temporal patterns, units, directionality
  \begin{itemize}[itemsep=2pt, parsep=0pt]
     \item[] - If the answer is generally correct but lacks details → 0.6–0.8.
     \item[] - If the answer has some errors but reasoning is reasonable → 0.3–0.5.
     \item[] - If the answer is mostly off-topic but mentions relevant terms → 0.1–0.2.
 \end{itemize}

\textbf{Return format reminder}:
\begin{itemize}[itemsep=2pt, parsep=0pt]
     \item[] - Output exactly: \{"answer": <score>\}
 \end{itemize}
   
\end{tcolorbox}
\captionof{table}{Details of the used prompts for the reasoning ability task evaluation}
\label{RAPromopt}}
\subsection{Time Series Generalization Ability Task Evaluation}
This section provides a detailed description of the implementation for cross-task generalization across forecasting, imputation, and multi-classification on LLADBench.

\subsubsection{Forecasting Tasks}
For the forecasting task, we employ the Mean Absolute Error (MAE) metric to quantify the discrepancy between the model-predicted numerical sequences and the ground truth values. This process comprises two primary stages: 1) parsing via pattern matching; and 2) leveraging LLM-assisted extraction to handle complex response paradigms. LLADBench performs formatting statistics and metric calculations by truncating sequences to a uniform length and filtering out outliers where the MAE exceeds 1000. Table~\ref{ForcstingPromopt} presents the prompts used for parsing mixed text-numerical inputs via LLMs. The script $Forcasting\_Test\_Evaluation\_Using\_LLM.py$ execution accepts two parameters:
\begin{itemize}[noitemsep, topsep=1pt] 
\item[]$--input\_dir$, which specifies the ``*generated\_predictions.jsonl'' file produced by Llama-factory, 
\item[]$--output\_dir$, which defines the directory where the formatted JSON output will be saved,
\item[] $--ground\_truth\_file$, it accepts the original ground truth values as input. 
\end{itemize}

{\begin{tcolorbox}[colback=white,colframe=black!70,title=Prompts of Parsing Forecasting Results,fonttitle=\bfseries,coltitle=white, boxrule=1pt]
\textbf{\color{black}Objective}:

You are a time series analysis expert.

\textbf{{\color{black}Objective}}:
 \begin{itemize}[itemsep=2pt, parsep=0pt]
     \item[] - Extract the predicted future time-series values from the student's answer text.
     \item[] - Return them strictly as a valid **Python list** (e.g., `[16.81, 18.09, 19.31, 20.25, 20.72]`).
     \item[] - If the student's answer **does not include explicit numeric predictions**, return an **empty list `[]`**.
 \end{itemize}

\textbf{{\color{black}Input example (student answer excerpt)}}:
 \begin{itemize}[itemsep=2pt, parsep=0pt]
     \item[] - example1: “Based on the given information and the trend observed in the time series, here are my predictions for the next 5 time series points:
16.81, 18.09, 19.31, 20.25, 20.72.” => should return [16.81, 18.09, 19.31, 20.25, 20.72]
     \item[] - example2: “let's generate the next 3 predictions:1. 1652 2. 1629 3. 1557” => should return [1652, 1629, 1557]
 \end{itemize}

\textbf{{\color{black}Output example (return only a Python list)}}:
 \begin{itemize}[itemsep=2pt, parsep=0pt]
     \item[] - [16.81, 18.09, 19.31, -20.25, 20.72]
 \end{itemize}

\textbf{Parsing \& Robustness requirements}:
  \begin{itemize}[itemsep=2pt, parsep=0pt]
     \item[] - Understand varied phrasings and formats (English/Chinese, newlines, comma/space-separated, numbers embedded in sentences, additional commentary).
     \item[] - **Return only the numeric list**; no extra text, units, keys, or explanations.
     \item[] - Accept integers/floats/scientific notation; parse correctly.
     \item[] - If multiple candidate sequences appear, return the set that is explicitly presented as “predictions”; if this cannot be uniquely determined, return `[]`.
     \item[] - If the answer provides ranges or uncertainty without explicit numbers, return `[]`.
 \end{itemize}

\end{tcolorbox}
\captionof{table}{Details of the used prompts for the generalization task evaluation in forecasting}
\label{ForcstingPromopt}}

\subsubsection{Imputation Tasks}
For the imputation task, we adopt a configuration similar to that of the forecasting task within the $Imputation\_Test\_Evaluation.py$ script. It first performs a coarse-grained string-list pattern matching, followed by LLM-based parsing to handle complex patterns that the scripted code fails to recognize. The corresponding prompts are detailed in Table~\ref{ImputationPromopt}.

{\begin{tcolorbox}[colback=white,colframe=black!70,title=Prompts of Parsing Forecasting Results,fonttitle=\bfseries,coltitle=white, boxrule=1pt]
\textbf{\color{black}Objective}:

You are a time series analysis expert.

\textbf{{\color{black}Objective}}:
 \begin{itemize}[itemsep=2pt, parsep=0pt]
     \item[] - Extract the predicted future time-series values from the student's answer text.
     \item[] - Return them strictly as a valid **Python list** (e.g., `[16.81, 18.09, 19.31, 20.25, 20.72]`).
     \item[] - If the student's answer **does not include explicit numeric predictions**, return an **empty list `[]`**.
 \end{itemize}

\textbf{{\color{black}Input example (student answer excerpt)}}:
 \begin{itemize}[itemsep=2pt, parsep=0pt]
     \item[] - example1: “Based on the given information and the trend observed in the time series, here are my predictions for the next 5 time series points:
16.81, 18.09, 19.31, 20.25, 20.72.” => should return [16.81, 18.09, 19.31, 20.25, 20.72]
     \item[] - example2: “let's generate the next 3 predictions:1. 1652 2. 1629 3. 1557” => should return [1652, 1629, 1557]
     \item[] - example3: "['X', 2012.0, 1511.0, 1182.0, 1223.0, 1917.0" => should return []
 \end{itemize}

\textbf{{\color{black}Output example (return only a Python list)}}:
 \begin{itemize}[itemsep=2pt, parsep=0pt]
     \item[] - [16.81, 18.09, 19.31, -20.25, 20.72] or []
 \end{itemize}

\textbf{Parsing \& Robustness requirements}:
  \begin{itemize}[itemsep=2pt, parsep=0pt]
     \item[] - Understand varied phrasings and formats (English/Chinese, newlines, comma/space-separated, numbers embedded in sentences, additional commentary).
     \item[] - **Return only the numeric list**; no extra text, units, keys, or explanations.
     \item[] - Accept integers/floats/scientific notation; parse correctly.
     \item[] - If multiple candidate sequences appear, return the set that is explicitly presented as “predictions”; if this cannot be uniquely determined, return `[]`.
     \item[] - If the answer provides ranges or uncertainty without explicit numbers, return `[]`.
     \item[] - For lists with incomplete closing brackets, you should still parse them.
     \item[] - Return an empty list for results where imputation failed.
 \end{itemize}

\end{tcolorbox}
\captionof{table}{Details of the used prompts for the generalization task evaluation in imputation}
\label{ImputationPromopt}}

\subsubsection{Multi-classification Tasks}
The multi-classification task involves complex text parsing, as the raw outputs often do not conform to a valid JSON format for practical applications. Since LLM exhibit robust performance in attribute matching and parsing, we evaluate the system based on the accuracy of the model-parsed results relative to the ground truth. The parsing script, $Classification\_Test\_Evaluation.py$, accepts three parameters:
\begin{itemize}[noitemsep, topsep=1pt] 
\item[]$--predicted\_file$, which specifies the ``*generated\_predictions.jsonl'' file by Llama-factory, 
\item[]$--output\_dir$, which defines the directory where the formatted JSON output will be saved,
\item[] $--ground\_truth\_file$, it accepts the original ground truth values as input. 
\end{itemize}

{\begin{tcolorbox}[colback=white,colframe=black!70,title=Prompts of Parsing Classification Results,fonttitle=\bfseries,coltitle=white, boxrule=1pt]
As a time series multi-class classification expert, assess if the student's answer matches the ground truth.

\textbf{{\color{black}Rules}}:
 \begin{itemize}[itemsep=2pt, parsep=0pt]
     \item[] - Match → 1.
     \item[] - Mismatch → 0.
 \end{itemize}

Output ONLY the score as a single integer (0 or 1). No explanations or extra text.

\textbf{{\color{black}Inputs}}:
 \begin{itemize}[itemsep=2pt, parsep=0pt]
     \item[] - Original question: Q.
     \item[] - Student answer: P.
     \item[] - Ground truth: Ground truth.
 \end{itemize}

\end{tcolorbox}
\captionof{table}{Details of the used prompts for the generalization task evaluation in multiple classification}
\label{ClassificationPromopt}}

\section{TKTO Optimization}\label{TKTO-SUP}
We present a comprehensive theoretical analysis of Time Series Kahneman-Tversky Optimization (TKTO), a reinforcement learning algorithm designed to enhance cross-task generalization in time series anomaly detection. Building upon the prospect theory framework of Kahneman and Tversky~\cite{kto}, TKTO extends the original KTO algorithm to the time series domain by introducing continuous preference scoring, multi-dimensional feedback integration, and soft-label regularization mechanisms.

\subsection{The Cross-Task Generalization Challenge}
MTSLLMs for time series AD face significant challenges in cross-task generalization. While SFT achieves strong performance on individual tasks, models often fail to transfer learned representations across different TS analysis tasks such as classification, forecasting, and imputation. Our motivation for incorporating reinforcement learning to bolster the model stems from the following considerations:
\textbf{1) Task-Specific Overfitting:} Models trained with standard cross-entropy loss tend to memorize task-specific patterns, limiting their ability to extract universal time series representations.
\textbf{2) Binary Feedback Limitations:} Traditional preference optimization methods (e.g., KTO) rely on binary good/bad labels, which are insufficient to capture the nuanced quality spectrum inherent in time series analysis outputs.
\textbf{3) Boundary Noise Sensitivity:} Hard decision boundaries in binary classification create training instability when samples lie near quality thresholds (e.g., 58-62 score range).
\textbf{4) Expensive Annotation Requirements:} Preference pair labeling for time series tasks requires domain expertise and is prohibitively expensive at scale.

\subsection{Prospect Theory Foundation}
Kahneman and Tversky's prospect theory formalizes how humans perceive outcomes relative to a reference point, exhibiting:
\textbf{1) Loss Aversion:} Humans are more sensitive to losses than equivalent gains, formally captured by $\lambda > 1$ in the value function.
\textbf{2) Diminishing Sensitivity:} The marginal value of gains/losses decreases as distance from the reference point increases, modeled by the concavity parameter $\alpha < 1$.
\textbf{3) Reference Dependence:} Utility is assessed relative to a reference point $z_0$ rather than absolute outcomes.
The canonical Kahneman-Tversky value function is~\cite{kto}:
\begin{equation}
v(z; \lambda, \alpha, z_0) = \begin{cases}
(z - z_0)^\alpha & \text{if } z \geq z_0 \text{ (gains)} \\
-\lambda (z_0 - z)^\alpha & \text{if } z < z_0 \text{ (losses)}
\end{cases}
\label{eq:kt_value}
\end{equation}
where typical empirical values are $\alpha \approx 0.88$ and $\lambda \approx 2.25$.

\subsection{From KTO to TKTO: Extending to Time Series}
TKTO adapts the KTO framework~\cite{kto} to address time series-specific requirements:
\textbf{1) Continuous Preference Modeling:} Replace binary labels $y \in \{0,1\}$ with continuous confidence scores $\alpha \in [0, 1]$ derived from multi-dimensional feedback.
\textbf{2) Multi-Task Feedback Integration:} Incorporate four evaluation dimensions: format compliance, task correctness, fluency, and human preference, with task correctness weighted highest for time series applications.
\textbf{3) Cross-Task Data Mixing:} Train on combined datasets spanning anomaly detection, classification, forecasting, and imputation to promote task-agnostic feature learning.
\textbf{4) Soft-Label Regularization:} Smooth preference boundaries to reduce noise and improve training stability in ambiguous quality regions.

\subsubsection{Core Loss Function Formulation of TKTO}
Let $\pi_\theta$ denote the ChatAD* model being optimized (with parameters $\theta$), and $\pi_{\text{ref}}$ the reference model (e.g., ChatAD after SFT). For a question-answer pair $(x, y)$ where $x$ represents the time series analysis question (including time series data, textual description) and $y$ is the generated answer, the TKTO loss is:
\begin{equation}
\mathcal{L}_{\text{TKTO}}(\pi_\theta) = \mathbb{E}_{(x,y,d) \sim \mathcal{D}} \left[ \alpha(d) \cdot \mathcal{L}_D(\pi_\theta, x, y) + (1 - \alpha(d)) \cdot \mathcal{L}_U(\pi_\theta, x, y) \right]
\label{eq:tkto_main}
\end{equation}
where: $\mathcal{D}$ is raining dataset containing Q \& A pairs and their multi-dimensional feedback scores $d \in \mathbb{R}^4$; $\alpha(d) \in [0, 1]$, continuous preference confidence score (key innovation of TKTO); $\mathcal{L}_D$ is desirable (preference) loss, encouraging high-quality outputs; and $\mathcal{L}_U$ is undesirable (rejection) loss, suppressing low-quality outputs.

\subsubsection{Desirable and Undesirable Loss Components}
The preference and rejection losses are defined using a logistic sigmoid function:

\begin{align}
\mathcal{L}_D(\pi_\theta, x, y) &= -\log \sigma(\beta \cdot \tau(x, y)) \label{eq:loss_d} \\
\mathcal{L}_U(\pi_\theta, x, y) &= -\log \sigma(-\beta \cdot \tau(x, y)) \label{eq:loss_u}
\end{align}

where $\sigma(z) = \frac{1}{1 + e^{-z}}$ is the sigmoid function, and:

\begin{equation}
\tau(x, y) = \log \frac{\pi_\theta(y|x)}{\pi_{\text{ref}}(y|x)}
\label{eq:tau}
\end{equation}

is the \textit{log-probability ratio} between the current model (ChatAD*) and the reference model (ChatAD). The hyperparameter $\beta > 0$ controls the strength of the KL-divergence constraint (typically $\beta \in [0.05, 0.2]$)~\cite{kto}.
When $\tau(x, y) > 0$: $\pi_\theta$ assigns higher probability to $y$ than $\pi_{\text{ref}}$, indicating the model favors this output more than the reference.
$\mathcal{L}_D$ rewards the model (low loss) when $\tau > 0$ for high-quality examples ($\alpha \to 1$).
$\mathcal{L}_U$ penalizes the model (high loss) when $\tau > 0$ for low-quality examples ($\alpha \to 0$).

\subsubsection{Continuous Preference Confidence Mechanism}
TKTO replaces binary preference labels with a continuous confidence score computed via a sigmoid transformation:

\begin{equation}
\alpha(d) = \sigma(\kappa \cdot (S(d) - \eta))
\label{eq:alpha}
\end{equation}
where, $S(d)$: Composite quality score computed from multi-dimensional feedback $d = [d_0, d_1, d_2, d_3]$; $\eta$: Threshold center (default: 60), determining the preference decision boundary; $\kappa > 0$: Steepness coefficient (default: 0.1), controlling the smoothness of the transition region.
Among Eq.~(\ref{eq:alpha}), the composite score $S(d)$ integrates both automated LLM feedback (LLMF) (in this study, we use the state-of-the-art GPT-5.1 model~\cite{openai_gpt5}) and human preference feedback (HPF):
\begin{equation}
S(d) = \epsilon \cdot \underbrace{(d_0 + d_1 + d_2)}_{\text{LLM Feedback (LLMF)}} + (1 - \epsilon) \cdot \underbrace{d_3}_{\text{Human Preference Feedback (HPF)}}
\label{eq:composite_score}
\end{equation}

where $\epsilon \in [0, 1]$ balances automated vs. human feedback (default: $\epsilon = 0.5$); $d_0 \in [0, 25]$: \textbf{Format Compliance} --- Evaluates whether the output adheres to structured requirements (e.g., JSON format, required fields); $d_1 \in [0, 50]$: \textbf{Time Series Task Correctness} --- Assesses accuracy of anomaly detection, prediction, or other time series analysis tasks (highest weight); $d_2 \in [0, 25]$: \textbf{Fluency} --- Measures natural language quality, grammatical correctness, and coherence
;$d_3 \in [0, 100]$: \textbf{Human Preference} --- Overall quality judgment from human annotators, serving as a holistic quality anchor.

Our design rationale for $d$ are as follows:
\textbf{1) Task Correctness Dominance ($d_1 = 50$):} Reflects the primacy of time series analysis accuracy over auxiliary attributes; \textbf{2) Auxiliary Dimensions ($d_0 + d_2 = 50$):} Format and fluency ensure outputs are usable and interpretable but do not dominate the objective; \textbf{3) Human Preference as Global Anchor ($d_3 = 100$):} Provides a comprehensive quality assessment that captures subtle aspects difficult to formalize in automated metrics.

\subsubsection{Prompts of LLM-Feedback Score}
For scores $\{d_0,  d_1, d_2\}$ are implemented using the GPT-5.1 model, with the corresponding prompts shown in Table~\ref{LLM-TKTO}.
{\begin{tcolorbox}[colback=white,colframe=black!70,title=Large model scoring prompts for TKTO optimization,fonttitle=\bfseries,coltitle=white, boxrule=1pt]
You are the highest-level expert in Time Series Task Evaluation, proficient in Forecasting, Classification, Imputation, Anomaly Detection, and Analysis.
Your task is to strictly and precisely evaluate a student model's answer (student) based on the true answer (true) and question (X). The evaluation must be delivered in three scores: Format, Correctness, and Fluency.

\textbf{1) Scoring Criteria:}
\begin{enumerate}[itemsep=3pt, parsep=0pt]
    \item {\color{blue}Format Compliance ($d_0$)}: [0 - 25 points].\\
    Requirement: Check if student adheres completely to the predetermined output format (e.g., closed arrays, correct JSON structure, single label).\\
    Zero Score: Any error causing automated processing failure (e.g., corrupted structure, extraneous text).
    \item {\color{blue}Task Correctness ($d_1$)}: [0 - 50 points].\\
    Requirement: Evaluate the mathematical or logical deviation between student and true.\\
    Forecasting/Imputation: Evaluate numerical deviation (e.g., RMSE < 10 is acceptable).\\
    Classification/Detection: Evaluate logical accuracy (matching true labels). This score reflects core task accuracy.
    \item {\color{blue}Fluency and Readability ($d_2$)}: [0 - 25 points].\\
    Requirement: Evaluate overall readability, organization, and adherence to human-like expressive habits. Deduct points for redundancy or unnecessary internal thought processes.
\end{enumerate}

\textbf{2) STRICT OUTPUT REQUIREMENT}:

You MUST and ONLY return the three scores as a precise Python list (List). The order must strictly be: Format Score ($d_0$), Correctness Score ($d_1$), Fluency Score ($d_2$).
\newline

\textbf{3) Examples}:
\begin{itemize}[itemsep=3pt, parsep=0pt]
    \item Perfect: [25.0, 48.5, 23.0]
    \item Error: [0.0, 15.2, 18.0]
\end{itemize}

You must perform a strict / harsh and fair evaluation, forbidding any form of laziness or ambiguous judgment. 
\end{tcolorbox}
\captionof{table}{Details of the used prompts for the LLM score
\label{LLM-TKTO}}

\section{Additional Results}~\label{ADDRES}

{
\small
\begin{table*}[htbp!]
\tabcolsep=0.08cm       % liejianju
\centering
\caption{Results of the mixed data studies for ChatAD-Llama3-8B, ChatAD-Qwen-2.5-7B, and ChatAD-Mistral-7B on TSEData (Warm-starting with the TimeMQA adapter). Values enclosed in parentheses denote the performance gain, calculated as ($Ours - Baseline$). Boldface highlights the best result.}
\resizebox{\linewidth}{!}{
\begin{tabular}{lccc|ccc|ccc}
\hline
\multirow{2}{*}{Model} & \multicolumn{3}{c|}{ChatAD-Llama3-8B} & \multicolumn{3}{c|}{ChatAD-Qwen-2.5-7B} & \multicolumn{3}{c}{ChatAD-Mistral-7B} \\
\cline{2-10}
& Acc $\uparrow$ ($\%$) & FPR $\downarrow$ ($\%$) & F1 $\uparrow$ ($\%$) & Acc $\uparrow$ ($\%$) & FPR $\downarrow$ ($\%$) & F1 $\uparrow$ ($\%$) & Acc $\uparrow$ ($\%$) & FPR $\downarrow$ ($\%$) & F1 $\uparrow$ ($\%$) \\
\hline
Base& 69.63 (+12.75) & 48.66 (-20.96) & 73.88 (+10.08) & 69.42 (+26.27) & 47.32 (-45.37) & 73.31 (+22.08) & 70.17 (+26.28) & 53.03 (-49.13) & 75.70 (+20.62) \\
Unmixed & 79.95 (+2.43) & 31.61 (-3.91) & 81.64 (+2.32) & 94.67 (+1.02) & 3.90 (-1.95) & 94.37 (+1.02) & 93.53 (+2.92) & 5.37 (-1.47) & 93.19 (+3.13) \\
Mixed & \textbf{82.38} & \textbf{27.70} & \textbf{83.96} & \textbf{95.69} & \textbf{1.95} & \textbf{95.39} & \textbf{96.45} & \textbf{3.90} & \textbf{96.32} \\
\hline
\end{tabular}
}
\begin{flushleft}

\end{flushleft}
\label{the-mixed-data}

\end{table*}
}
\end{document}